\documentclass[10pt,twocolumn,letterpaper]{article}

\usepackage{acpr}
\usepackage{times}
\usepackage{epsfig}
\usepackage{graphicx}
\usepackage{amsmath}
\usepackage{amssymb}
\usepackage[bottom]{footmisc}

\usepackage{url}
\usepackage{subcaption}
\usepackage{float}

\usepackage{algorithm}
\usepackage{algorithmicx}
\usepackage{algpseudocode}
\algnewcommand{\LineComment}[1]{\State \(\triangleright\) #1}

\usepackage{tabularx}

\usepackage{xcolor,colortbl}
\definecolor{Gray}{gray}{0.85}
\definecolor{LightCyan}{rgb}{0.88,1,1}
\definecolor{LightGreen}{rgb}{0.2,0.8,0.2}

\newcommand\circlednumber[1]{\raisebox{.5pt}{\textcircled{\raisebox{-.9pt} {#1}}}}

\newcommand\fauxsc[1]{\fauxschelper#1 \relax\relax}
\def\fauxschelper#1 #2\relax{%
  \fauxschelphelp#1\relax\relax%
  \if\relax#2\relax\else\ \fauxschelper#2\relax\fi%
}
\def\Hscale{.80}\def\Vscale{.72}\def\Cscale{1.0}
\def\fauxschelphelp#1#2\relax{%
  \ifnum`#1>``\ifnum`#1<`\{\scalebox{\Hscale}[\Vscale]{\uppercase{#1}}\else%
    \scalebox{\Cscale}[1]{#1}\fi\else\scalebox{\Cscale}[1]{#1}\fi%
  \ifx\relax#2\relax\else\fauxschelphelp#2\relax\fi}
	
\newcommand{\argmin}{\arg\!\min}

\usepackage[pagebackref=true,breaklinks=true,letterpaper=true,colorlinks,bookmarks=false]{hyperref}

\acprfinalcopy 

\ifacprfinal\fi
\begin{document}

\title{Analyzing structural characteristics of object category representations from their semantic-part distributions}

\author{Ravi Kiran Sarvadevabhatla\\
Video Analytics Lab\\
Supercomputer Education and Research Centre\\
Indian Institute of Science, Bangalore, India \\
{\tt\small ravikiran@ssl.serc.iisc.in}\\
\and
R. Venkatesh Babu\\
Video Analytics Lab\\
Supercomputer Education and Research Centre\\
Indian Institute of Science, Bangalore, India \\
{\tt\small venky@serc.iisc.in}
}

\maketitle

\begin{abstract}
Studies from neuroscience show that part-mapping computations are employed by human visual system in the process of object recognition. In this work, we present an approach for analyzing semantic-part characteristics of object category representations. For our experiments, we use category-epitome, a recently proposed sketch-based spatial representation for objects. To enable part-importance analysis, we first obtain semantic-part annotations of hand-drawn sketches originally used to construct the corresponding epitomes.  We then examine the extent to which the semantic-parts are present in the epitomes of a category and visualize the relative importance of parts as a word cloud. Finally, we show how such word cloud visualizations provide an intuitive understanding of category-level structural trends that exist in the category-epitome object representations. 
\end{abstract}

\section{Introduction}

Studies from neuroscience show that structural part-mapping computations are employed by the human visual system in the process of recognition~\cite{Lovett2009216}. Put another way, the presence of certain parts seems to be anticipated by the visual system when it attempts to recognize an object. The knowledge of what these parts are and their relative importance for the overall task of recognition can lead to insights regarding the neuro-visual representation of objects.

\begin{figure}[t]
\centering
\includegraphics[width=.48\textwidth]{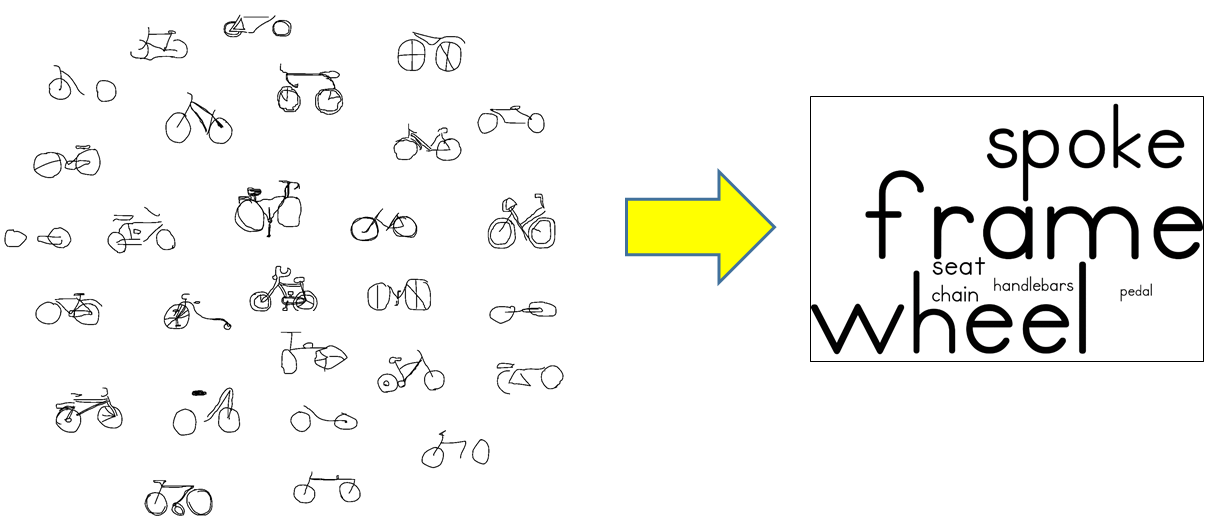}
\caption{Sparsified yet recognizable freehand sketch representations of object category \texttt{bicycle} on the left. Examining such a large number of instances visually for structural similarities can be ineffective. Instead, the approach we propose captures the structurally significant parts as a semantic-part word cloud (on the right). The size of a part's name in the word cloud reflects its importance across the set of sparsified representations of the category.}
\label{fig:epitomestowordle}
\end{figure}

\begin{figure*}[t]
	\centering
  \includegraphics[width=0.9\textwidth,height=0.25\linewidth]{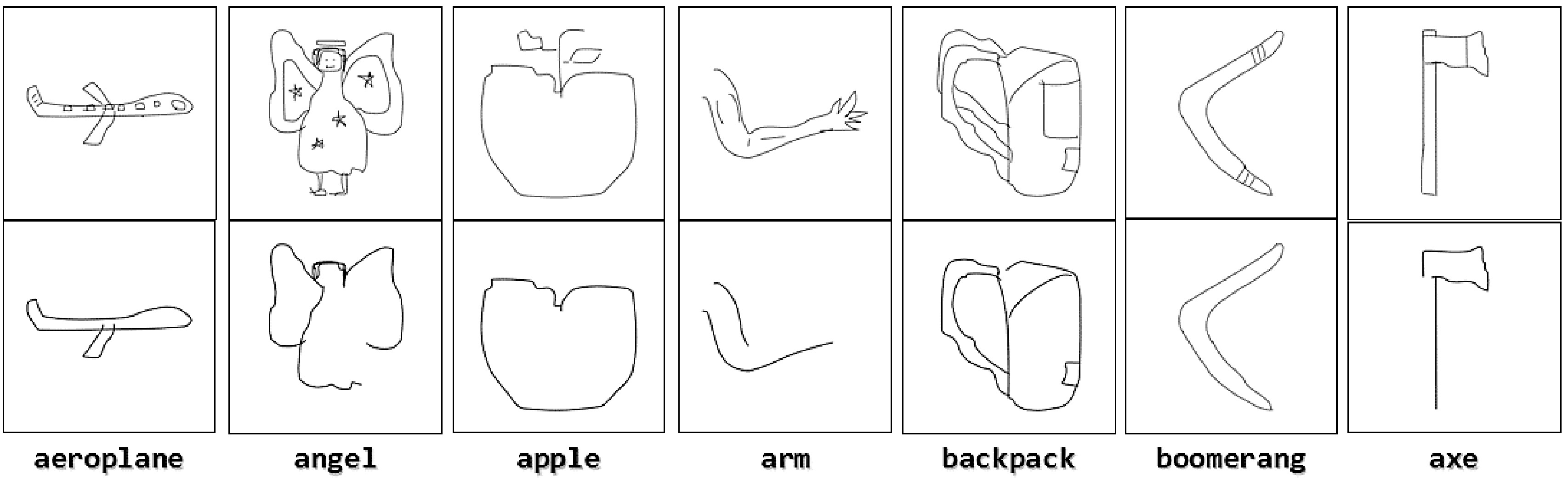}
	\caption{Original sketches (top row) and corresponding \textit{category-epitome}s (bottom row) for various object categories. Figure has been taken from ~\cite{eotd}.}
	\label{fig:epitomes}
\end{figure*}

In a recent work, Sarvadevabhatla et al.~\cite{eotd} describe the construction of sketch-based spatial representations for object categories termed \textit{category-epitome}s. The epitomes are constructed to as sparse as possible while still being machine-recognizable (see Figure ~\ref{fig:epitomes}). To study these epitomes, one possibility would be to visually examine them for structural similarities on a per-category basis. However, if the number of such epitomes is large, visual examination can be ineffective. An alternate approach would be to examine the distribution of semantic-parts\footnote{E.g. spokes, seat, wheel, handle etc. are the semantic parts of a bicycle. We use the term semantic-parts to distinguish from the common interpretation of an object part as a certain spatial, unnamed portion of an object.} in the epitomes of each category. As we show in this work, such an approach can lead to an intuitive understanding of category-specific ``signature" structural elements (parts) which persist in \textit{category-epitome}s (see Figure \ref{fig:epitomestowordle}). Moreover, the \textit{category-epitome}s we study have been obtained using human-drawn sketches as a starting point. Therefore, our approach also creates the possibility of analyzing the underlying human neuro-visual representations as well.

\section{Related Work}
\label{sec:relatedwork}

Determining the relative importance of part-level structural primitives for object category understanding has been explored only to a limited extent. Guo et al.~\cite{Guo:2014:SRM:2628615.2628633} present an importance measure of shape parts based on their ability to reconstruct the whole object shape of 2-D silhouettes. However, the authors interpret parts to mean segments on the contour of the object. Ma et al.~\cite{Ma} propose a perception-based method to segment a sketch into semantically meaningful parts. Interestingly, they demonstrate the effectiveness of utilizing semantic parts rather than just consider parts as unnamed ``regions" of the object. To the best of our knowledge, the relative importance of the semantic parts has not been studied.

\section{Construction of a part-annotated sketch database}
\label{sec:annotation}

As the first step towards the semantic-part based understanding of \textit{category-epitome}s, we manually annotated hand-drawn sketches from $13$ categories\footnote{\texttt{airplane, bicycle, bus, car (sedan), cat, cow, dog, flying bird, horse, person walking, potted plant, sheep, train}} from the sketch database of Eitz et al.~\cite{eitz} for our analysis. 
A direction of research we intend to pursue in future involves simultaneous analysis of image and sketch based categories (whose part-level segmentations have been provided). With this in mind, the $13$ categories we examine were chosen to overlap with PASCAL-parts~\cite{pascalparts} -- an image dataset containing part-level segmentations of $20$ object categories. From the $20$ categories in PASCAL-parts, we retained only those containing at least two dominant labeled parts. For example, category \texttt{tv} had only one dominant labeled part (\textit{screen}) and therefore was not admissible. Within the sketches of a category, we considered only the correctly classified sketches since \textit{category-epitome}s, by definition, cannot be constructed for misclassified sketches. Please refer to Section $4$ of ~\cite{eotd} for details regarding \textit{category-epitome} construction. 

\begin{figure}[t]
\centering
\includegraphics[width=.48\textwidth]{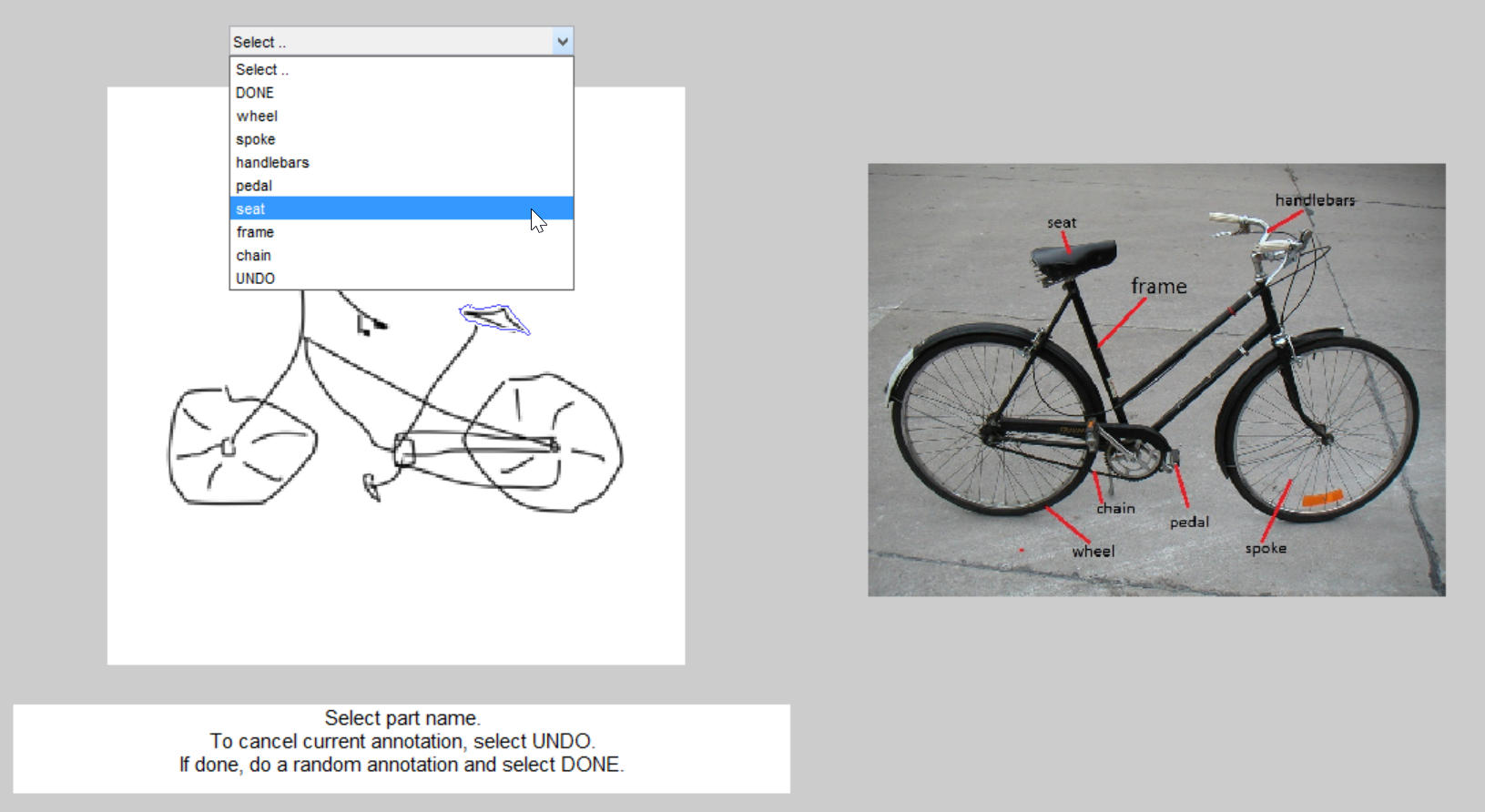}
\caption{Screenshot of our annotation system. The sketch to be annotated and the list of parts can be seen towards the left side. The reference image to guide the annotators for part names and locations is on the right.}
\label{fig:screenshot}
\end{figure}

\begin{figure}[t]
\centering
\includegraphics[width=.48\textwidth]{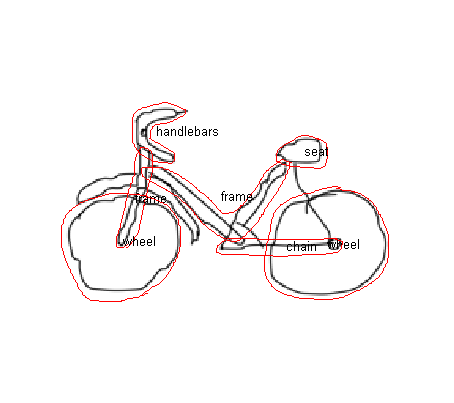}
\caption{User annotated sketch from category \texttt{bicycle}}
\label{fig:sampleannot}
\end{figure}

The annotation of part contours in the sketches was performed by $10$ annotators who used an annotation tool developed in-house (see Figure \ref{fig:screenshot}). In the figure, the sketch to be annotated is on the left. As an annotation guide, a prototypical image of the category, labeled with parts, was also provided alongside the sketch. The annotators were provided basic guidelines on annotation to ensure reasonably compact boundary contours enclosing each semantic-part. At the end, we obtained semantic-part contour annotations for $283$ sketches across $13$ object categories for an average of $22$ sketches per category. A sample annotation can be viewed in Figure \ref{fig:sampleannot}. For each annotated sketch, the corresponding sparsified representation (\textit{category-epitome}) was obtained from the epitome data provided by ~\cite{eotd} at \url{http://val.serc.iisc.ernet.in/eotd/epitome_images/}. 

\section{Our approach}
\label{sec:overview}

\begin{figure*}[!htbp]
    \centering    
				\includegraphics[width=0.8\textwidth,height=0.25\linewidth]{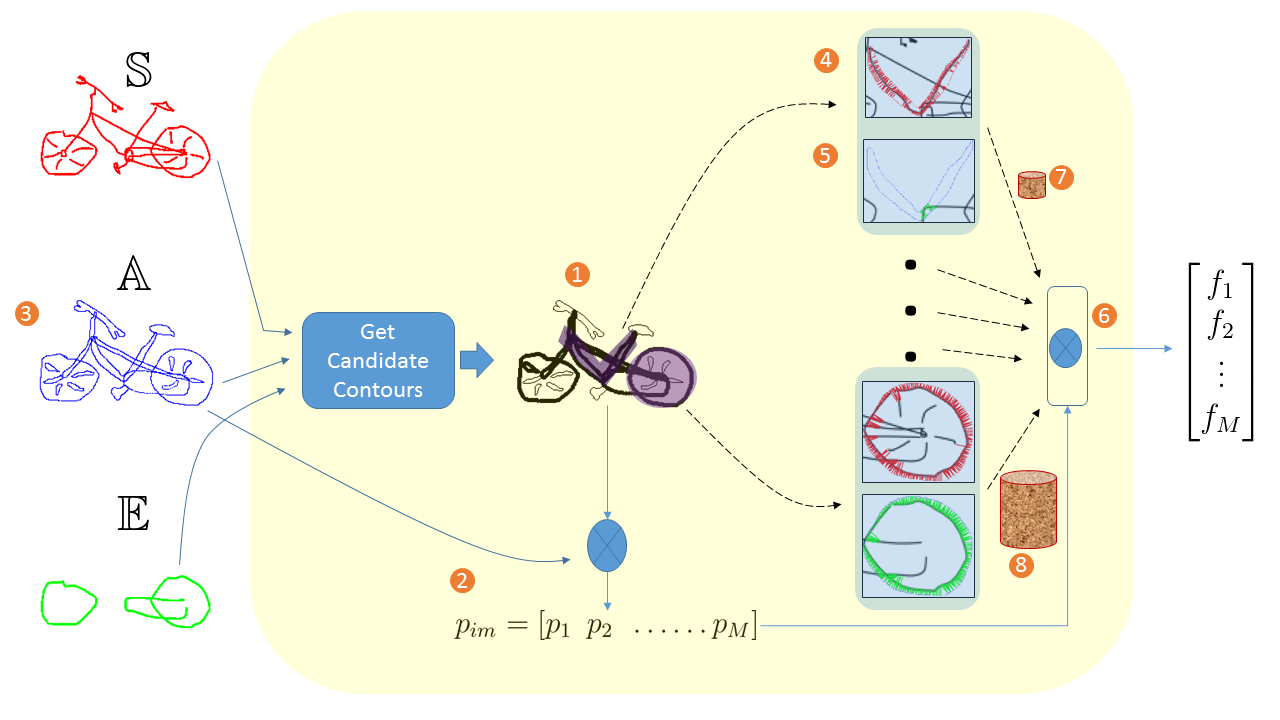}		
						\caption{Determining part-importances given the full sketch $\mathbb{S}$, the corresponding \textit{category-epitome} $\mathbb{E}$ and the set of user annotations $\mathbb{A}$ from the category \texttt{bicycle}. The output is a vector of semantic-part importances. Locations with numbered orange circles indicate key aspects of the pipeline. The sketches and annotation data have been color-coded for visualization purposes. This figure is best viewed in color.}
						\label{fig:overview}    
	\end{figure*}

An overview of our approach can be seen in Figure \ref{fig:overview}. In the figure, locations with numbers circled in orange correspond to important processing stages which we shall refer to in the discussion that follows. We utilize an example from the category \texttt{bicycle} for the purpose of illustration. Let $\mathbb{S}$ be the original full-sketch image from category $\mathcal{C}$, $\mathbb{E}$ its \textit{category-epitome} and $\mathbb{A}$, the set of 2-D contour points which correspond to user-annotated boundaries of semantic-parts in $\mathbb{S}$. Since the \textit{category-epitome} is constructed from its full-sketch counterpart, we have $\mathbb{E} \subseteq \mathbb{S}$ (i.e. the set of sketch strokes in the epitome is a subset of the strokes in the corresponding full-sketch). Let us also suppose the cardinality of $\mathbb{A}$ is $n_a$ and the number of semantic-parts in $\mathcal{C}$ is $M$.

\subsection{Obtaining candidate part contours}
\label{sec:candidate}

In some instances, part contours may enclose an insignificant number of pixels from epitome $\mathbb{E}$. As the first step, we filter out such contours. For each part contour, we compute the number of stroke pixels $n_{full}$ in $\mathbb{S}$ that lie within the part's boundary. We also compute the number of stroke pixels $n_{epi}$ in $\mathbb{E}$ that lie within the part's boundary. If the ratio $\frac{n_{epi}}{n_{full}}$ is larger than a threshold, the contour is added to the candidate contour list. In Figure \ref{fig:overview}, the candidate contours are shown in bold (region labeled \circlednumber{1}). Note that multiple occurrences of the same semantic-part type (e.g. spokes of the \texttt{bicycle}) which satisfy the threshold criterion are counted independently. To avoid undue importance to multiple occurrences of the same part type, we normalize by the corresponding part contour counts in full sketch to obtain a `coarse' part-importance factor $p_{im}$ for each semantic-part (see  \circlednumber{2} in Figure \ref{fig:overview}). We term it a `coarse' importance factor since it implicitly takes raw pixel counts into account for determining part importance. Shortly, we shall see how the spatial structure of the stroke is also utilized in determining the final semantic-part importance.

\begin{figure*}[!ht]
\caption{Importance of semantic structural parts for object categories : Each image shows a word cloud of parts for epitomes of each category. The size of the part name indicates its relative importance across epitomes of the category. The above depictions are for \textsc{Length} stroke sequence ordering.}	
    \centering
    \begin{minipage}{0.18\linewidth}
        \centering        
				\frame{\includegraphics[width=0.9\linewidth]{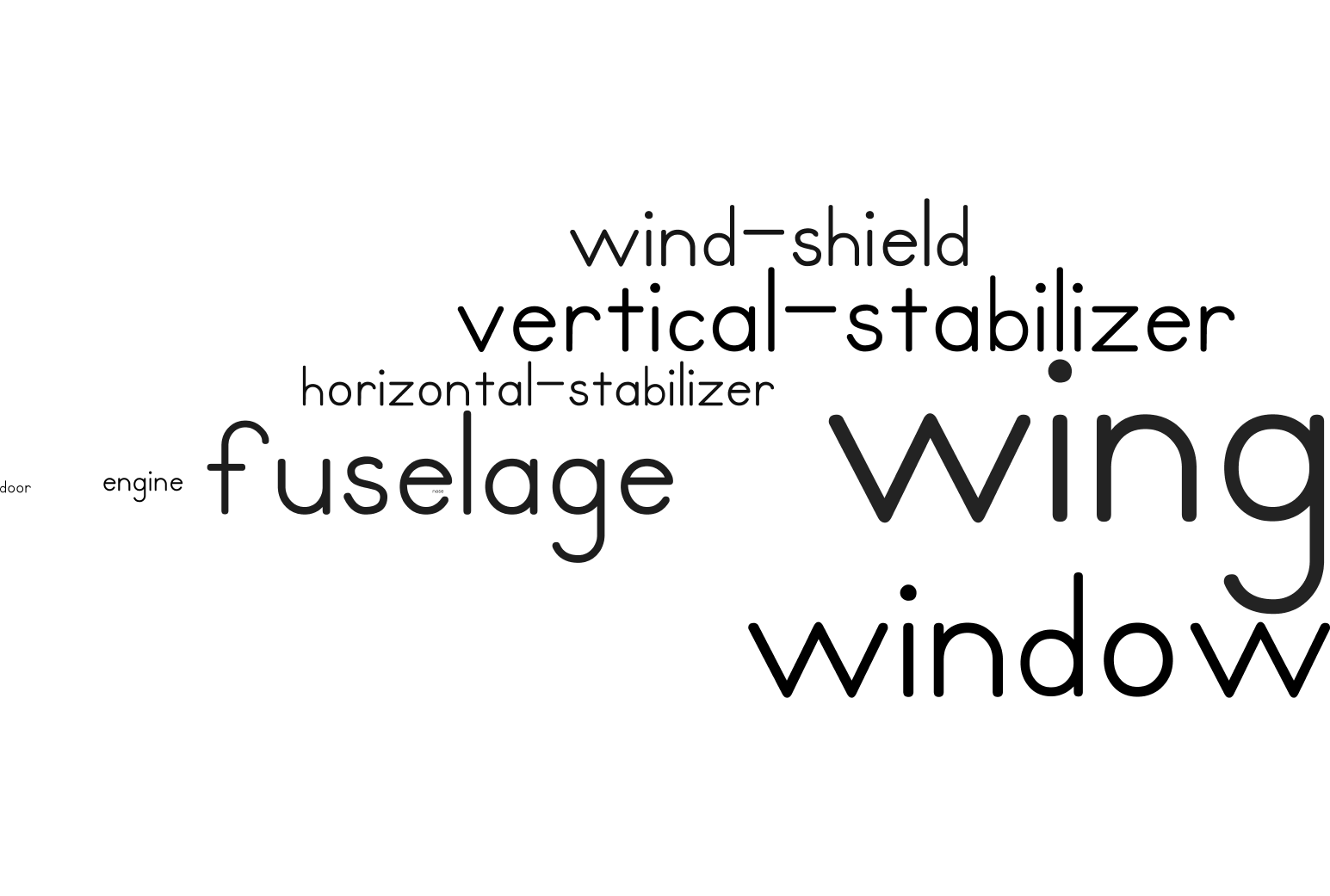}}   
						\caption*{\texttt{airplane}}
    \end{minipage}   
		\begin{minipage}{0.18\linewidth}
        \centering        
				\frame{\includegraphics[width=0.9\linewidth]{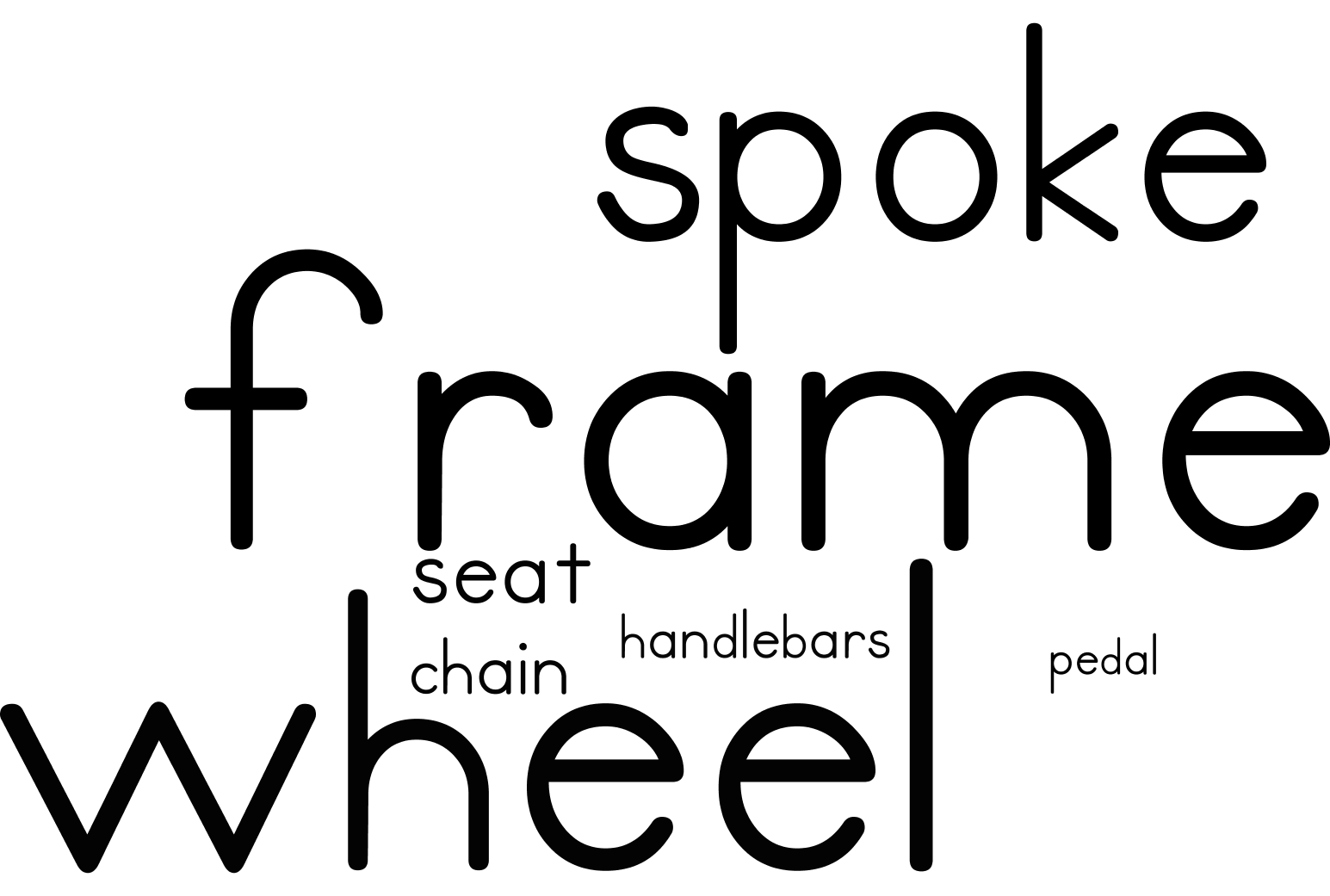}}     
					\caption*{\texttt{bicycle}}
    \end{minipage} 
		\begin{minipage}{0.18\linewidth}
        \centering        
				\frame{\includegraphics[width=0.9\linewidth]{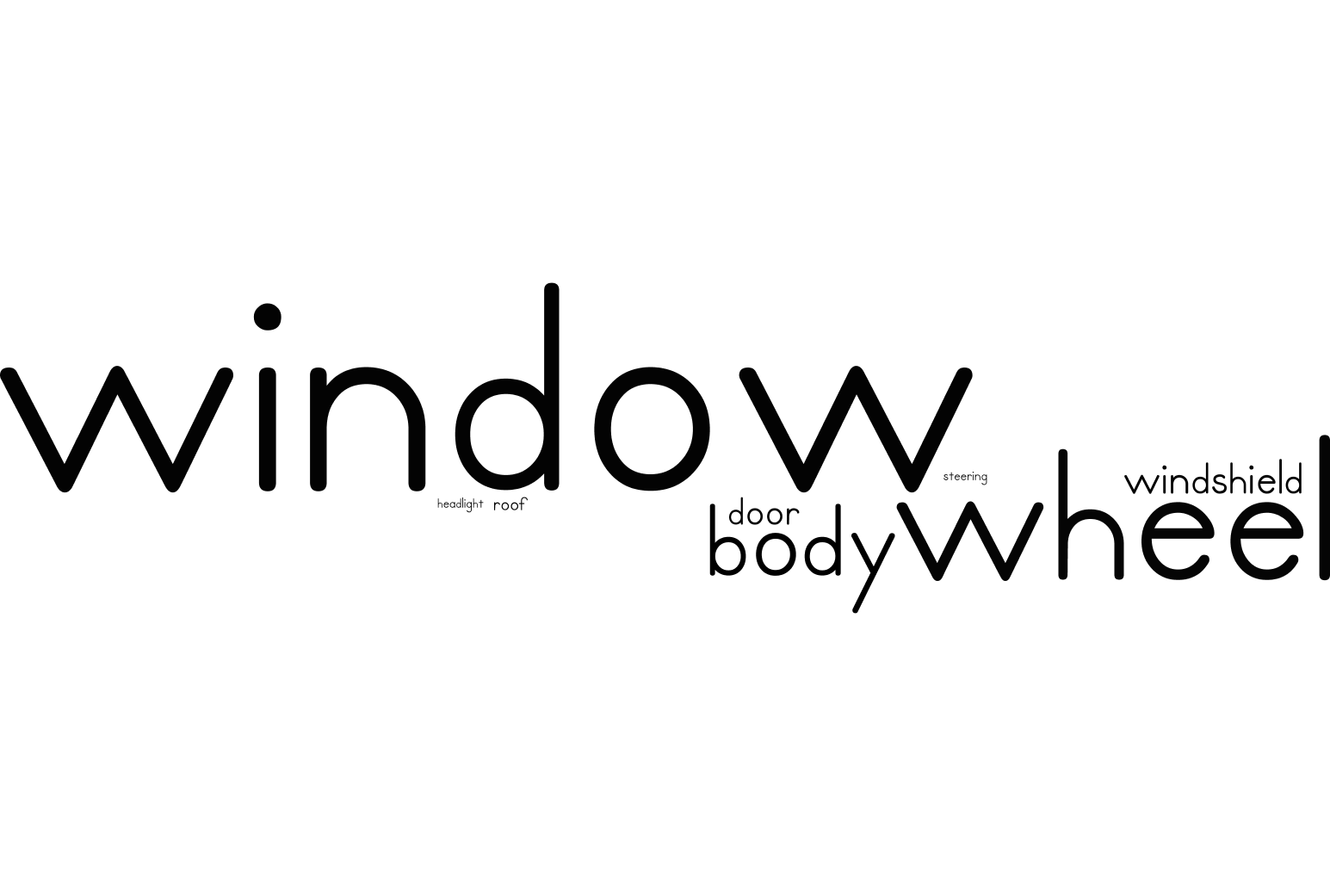}}     
					\caption*{\texttt{bus}}
    \end{minipage} 
		\begin{minipage}{0.18\linewidth}
        \centering        
				\frame{\includegraphics[width=0.9\linewidth]{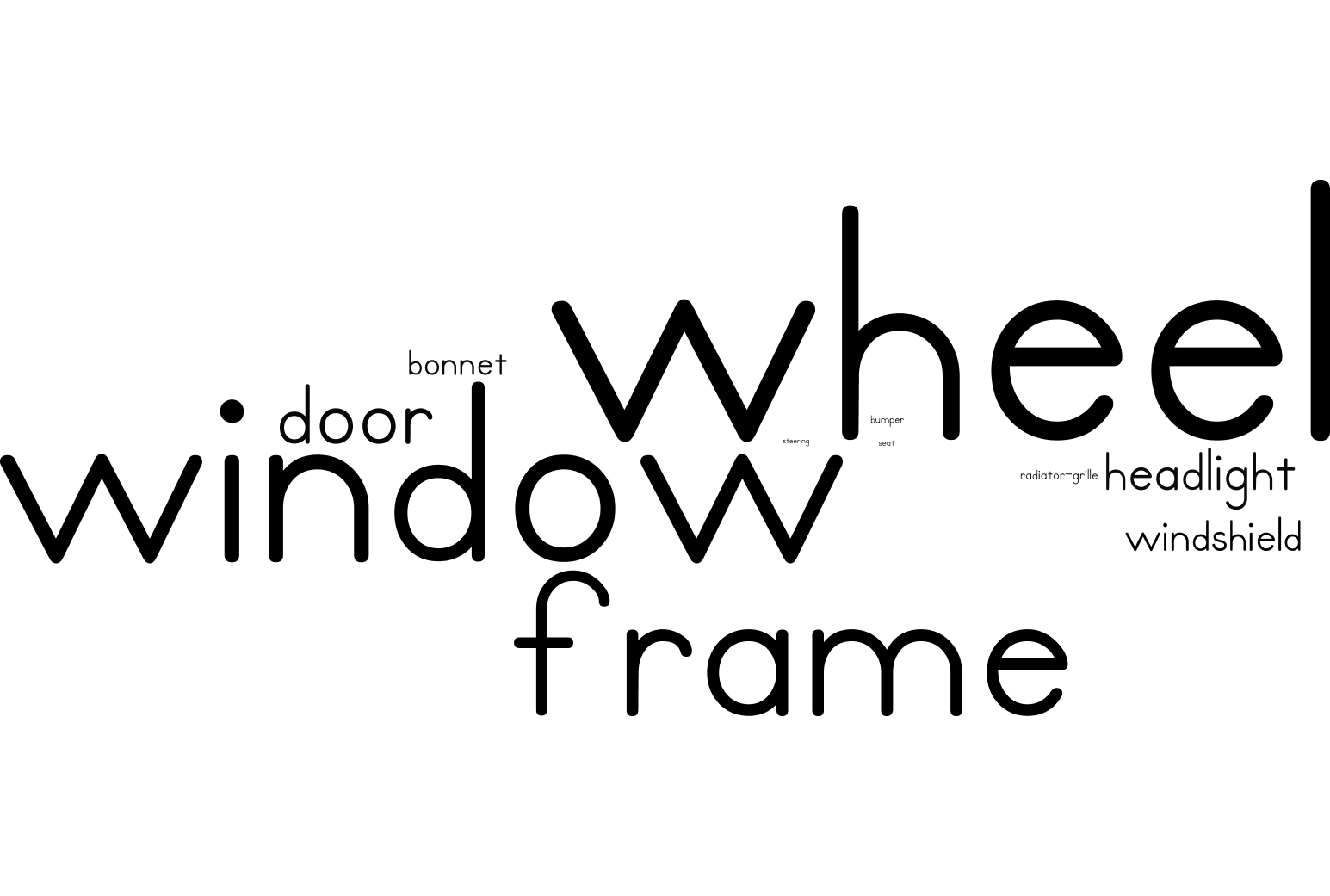} } 
						\caption*{\texttt{car}}
    \end{minipage} 
		\vspace{5mm}
		\begin{minipage}{0.18\linewidth}
        \centering        
				\frame{\includegraphics[width=0.9\linewidth]{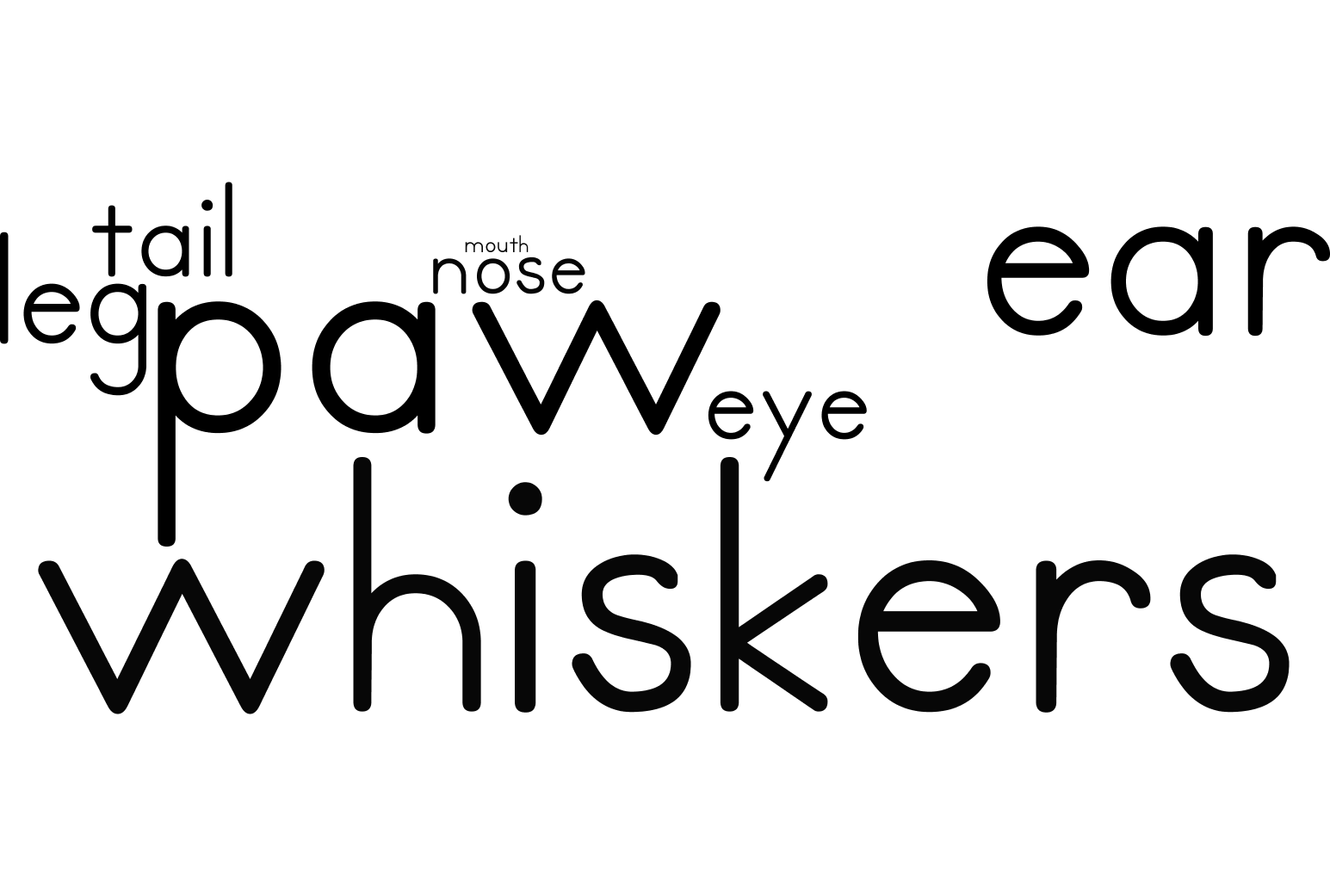} } 
						\caption*{\texttt{cat}}
    \end{minipage}     
		\begin{minipage}{0.18\linewidth}
        \centering        
				\frame{\includegraphics[width=0.9\linewidth]{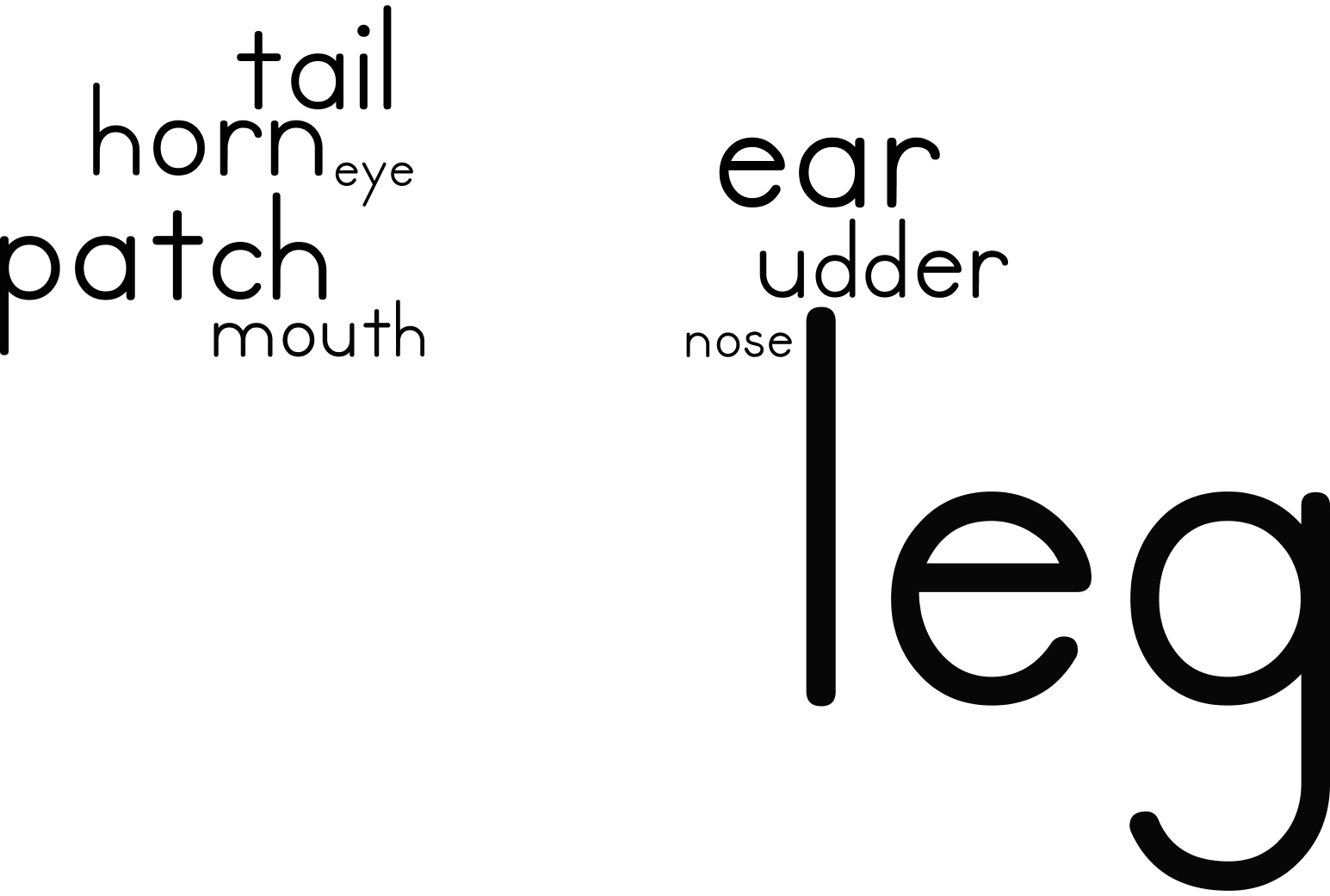}  }  
				\caption*{\texttt{cow}}
    \end{minipage}
		\begin{minipage}{0.18\linewidth}
        \centering        
				\frame{\includegraphics[width=0.9\linewidth]{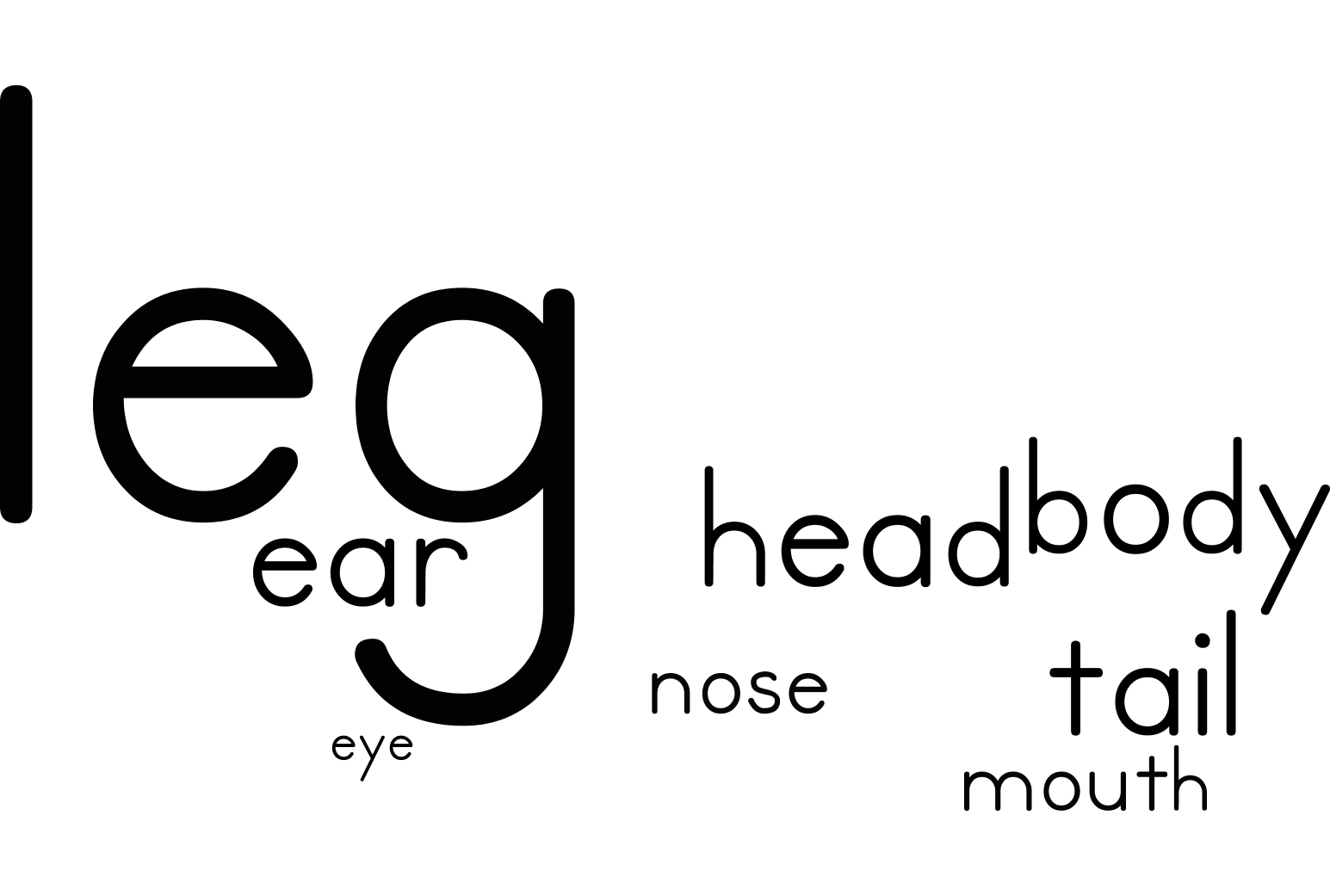}  }  
				\caption*{\texttt{dog}}
    \end{minipage} 
		\begin{minipage}{0.18\linewidth}
        \centering        
				\frame{\includegraphics[width=0.9\linewidth]{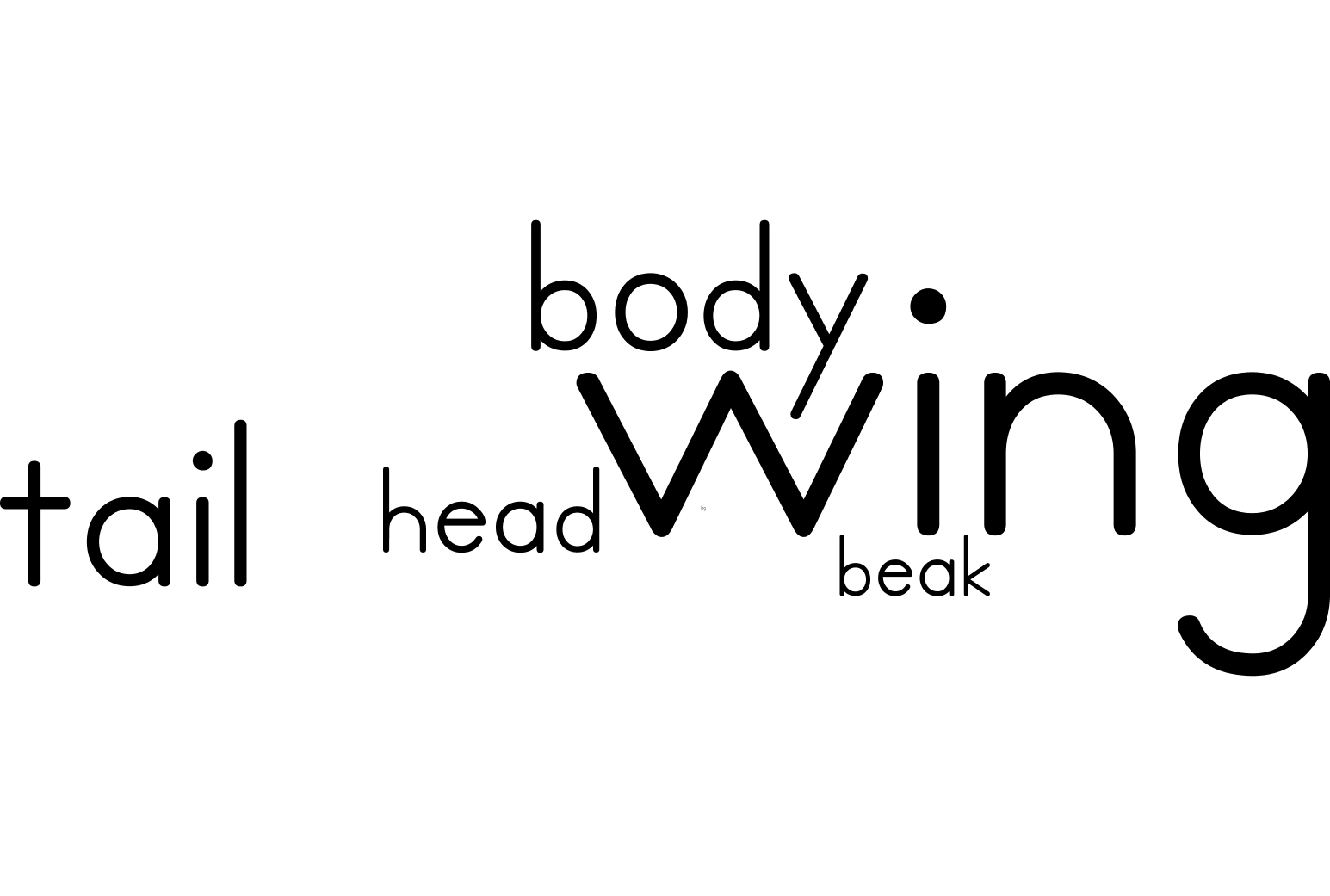}}       
				\caption*{\texttt{flying bird}}
    \end{minipage}  
		\begin{minipage}{0.18\linewidth}
        \centering        
				\frame{\includegraphics[width=0.9\linewidth]{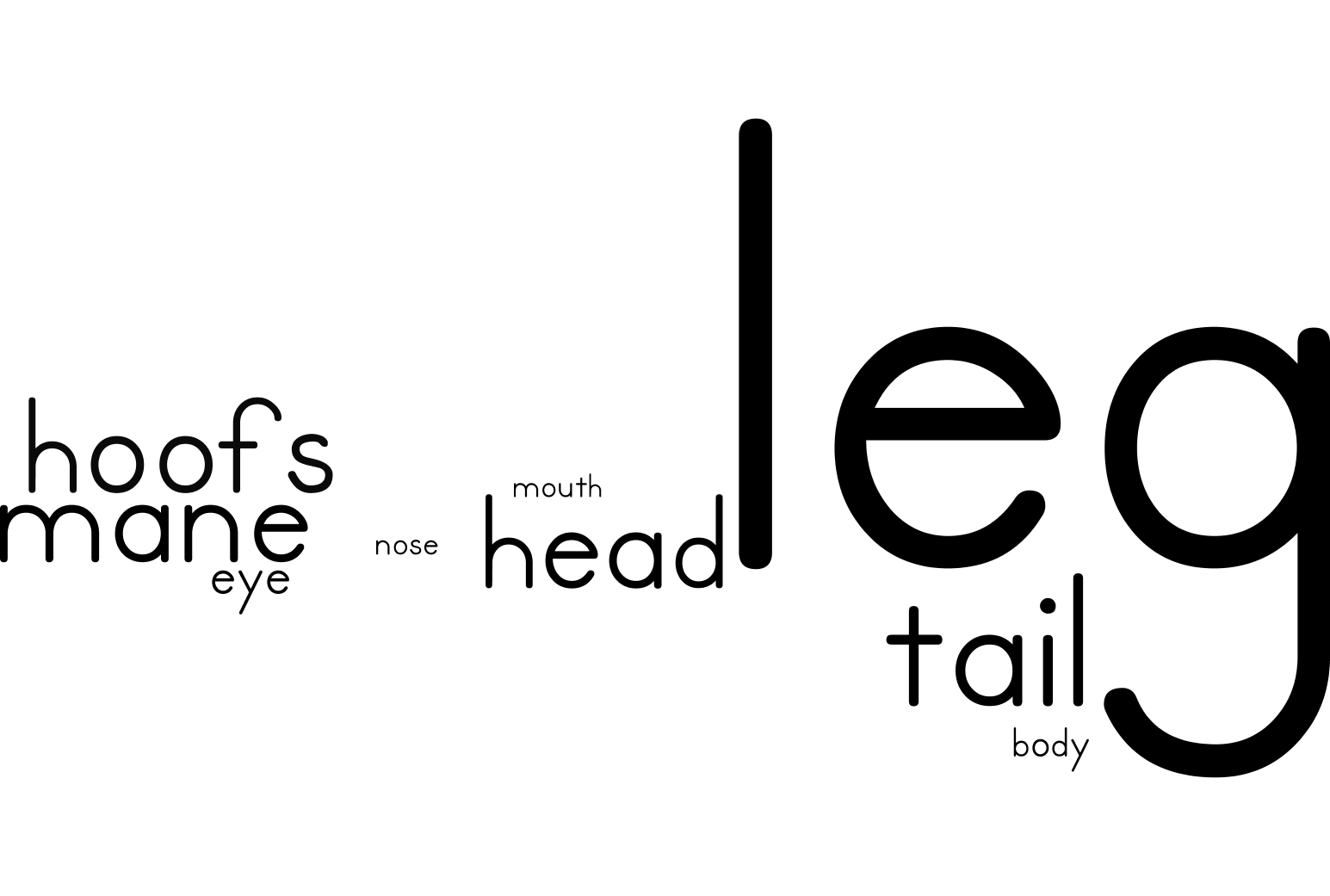}}       
				\caption*{\texttt{horse}}
    \end{minipage} 
		\begin{minipage}{0.18\linewidth}
        \centering        
				\frame{\includegraphics[width=0.9\linewidth]{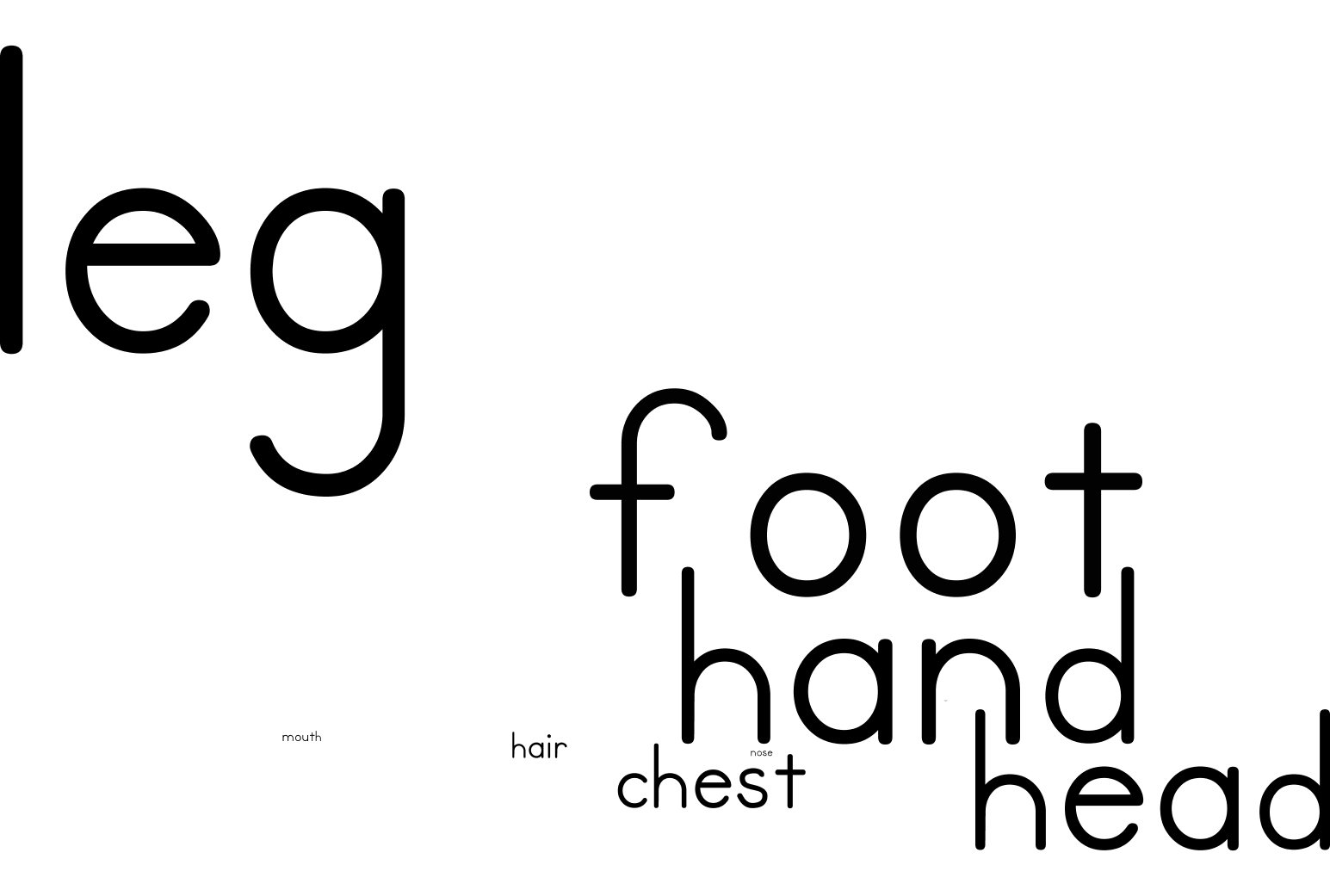}}       
				\caption*{\texttt{person walking}}
    \end{minipage} 
		\vspace{5mm}
    \centering			
		\begin{minipage}{0.18\linewidth}
        \centering        
				\frame{\includegraphics[width=0.9\linewidth]{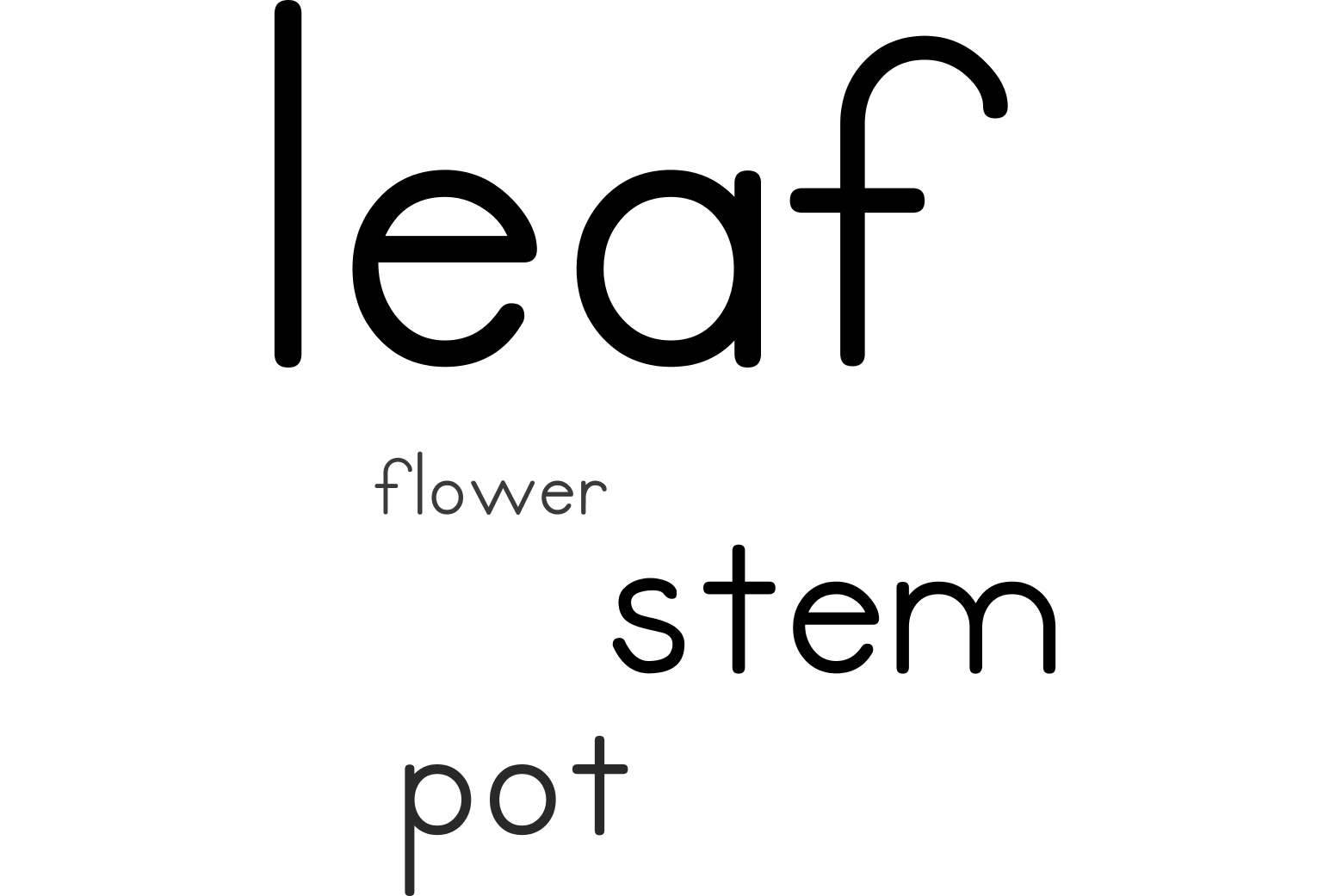}}       
				\caption*{\texttt{potted plant}}
    \end{minipage} 
		\begin{minipage}{0.18\linewidth}
        \centering        
				\frame{\includegraphics[width=0.9\linewidth]{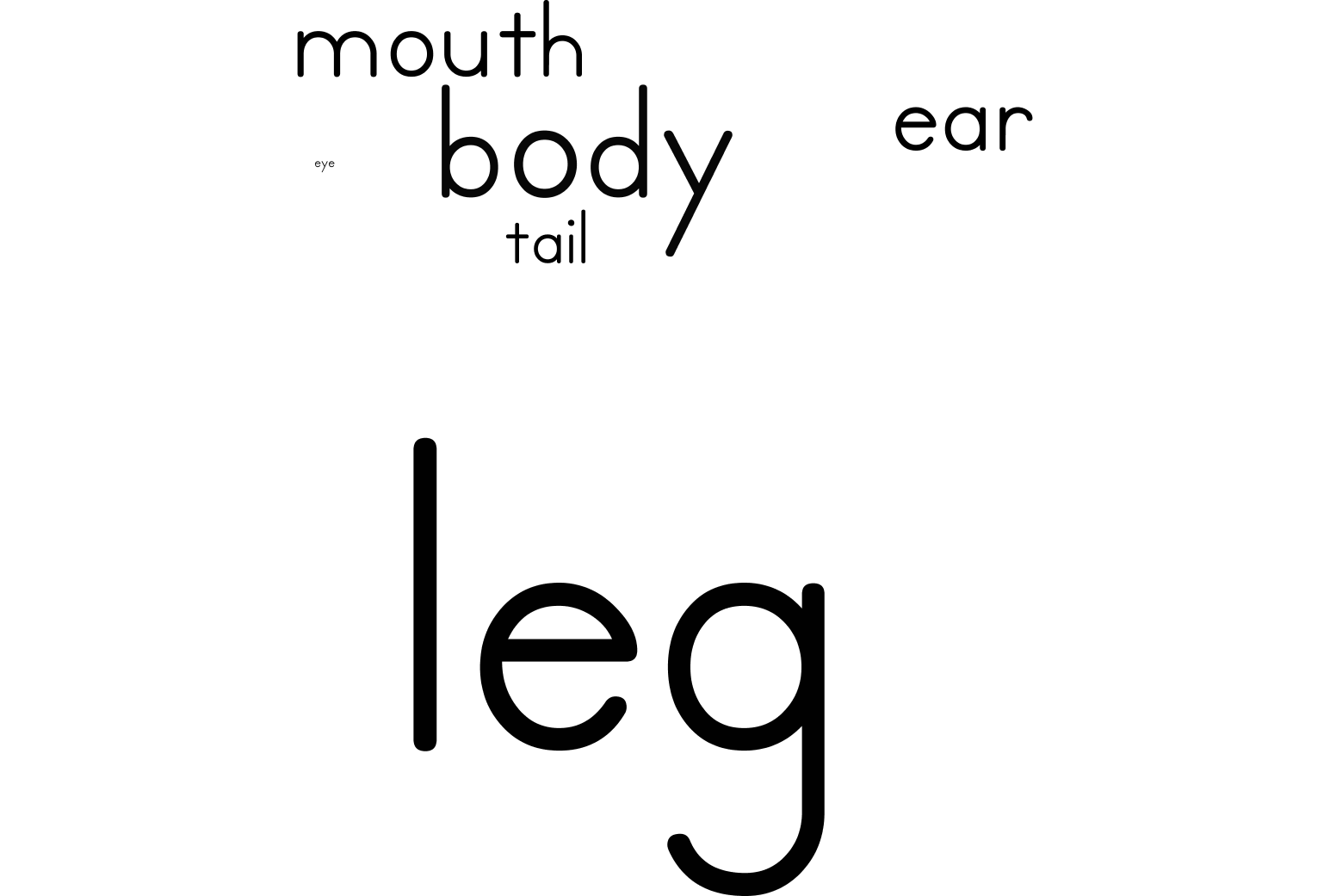}}       
				\caption*{\texttt{sheep}}
    \end{minipage} 		
		\begin{minipage}{0.18\linewidth}
        \centering        
				\frame{\includegraphics[width=0.9\linewidth]{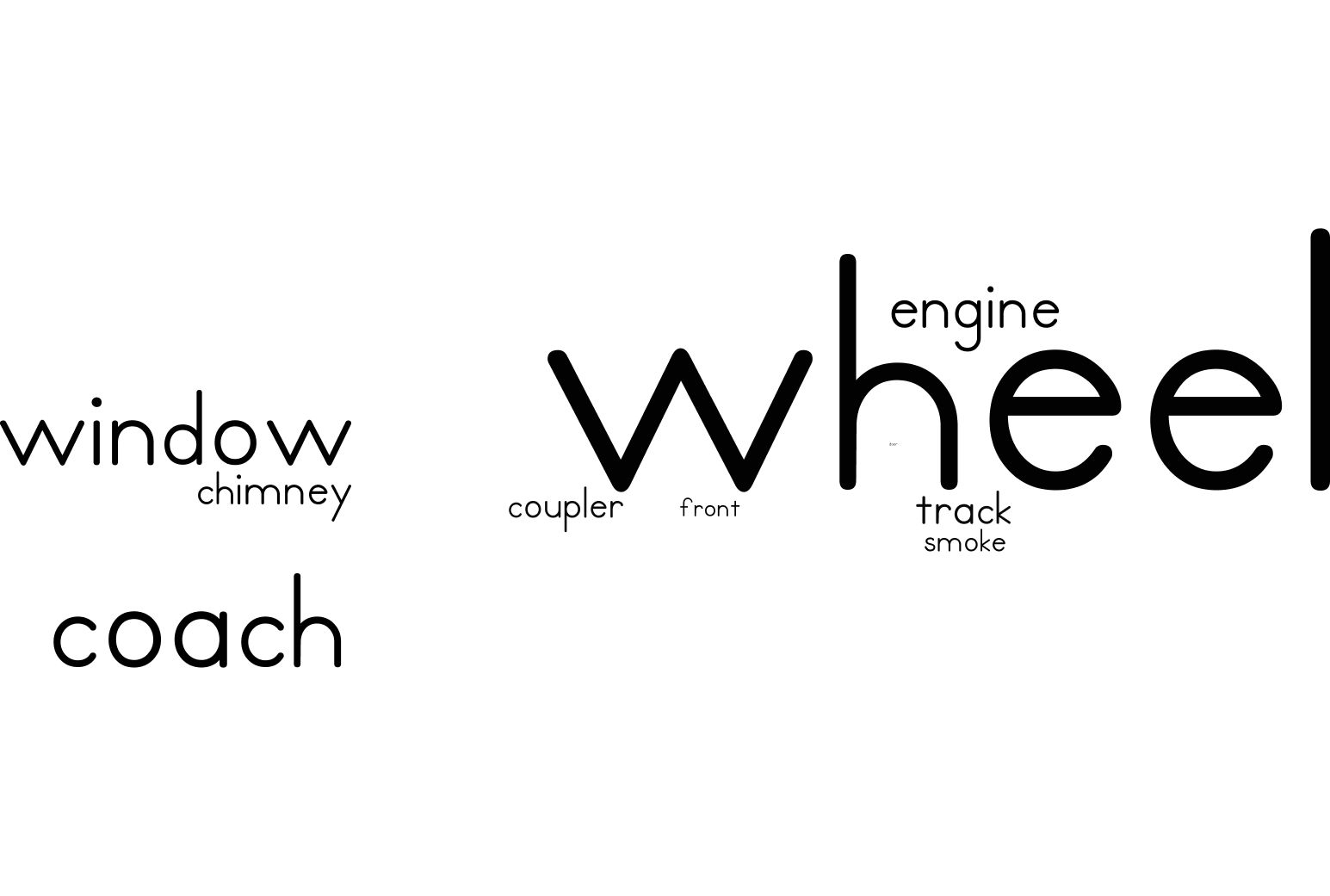}}       
				\caption*{\texttt{train}}
    \end{minipage} 				
\label{fig:epitome-wordle}
\end{figure*}

\begin{figure*}[!ht]
    \centering		
		\begin{minipage}{0.18\linewidth}
        \centering        
				\frame{\includegraphics[width=0.9\linewidth]{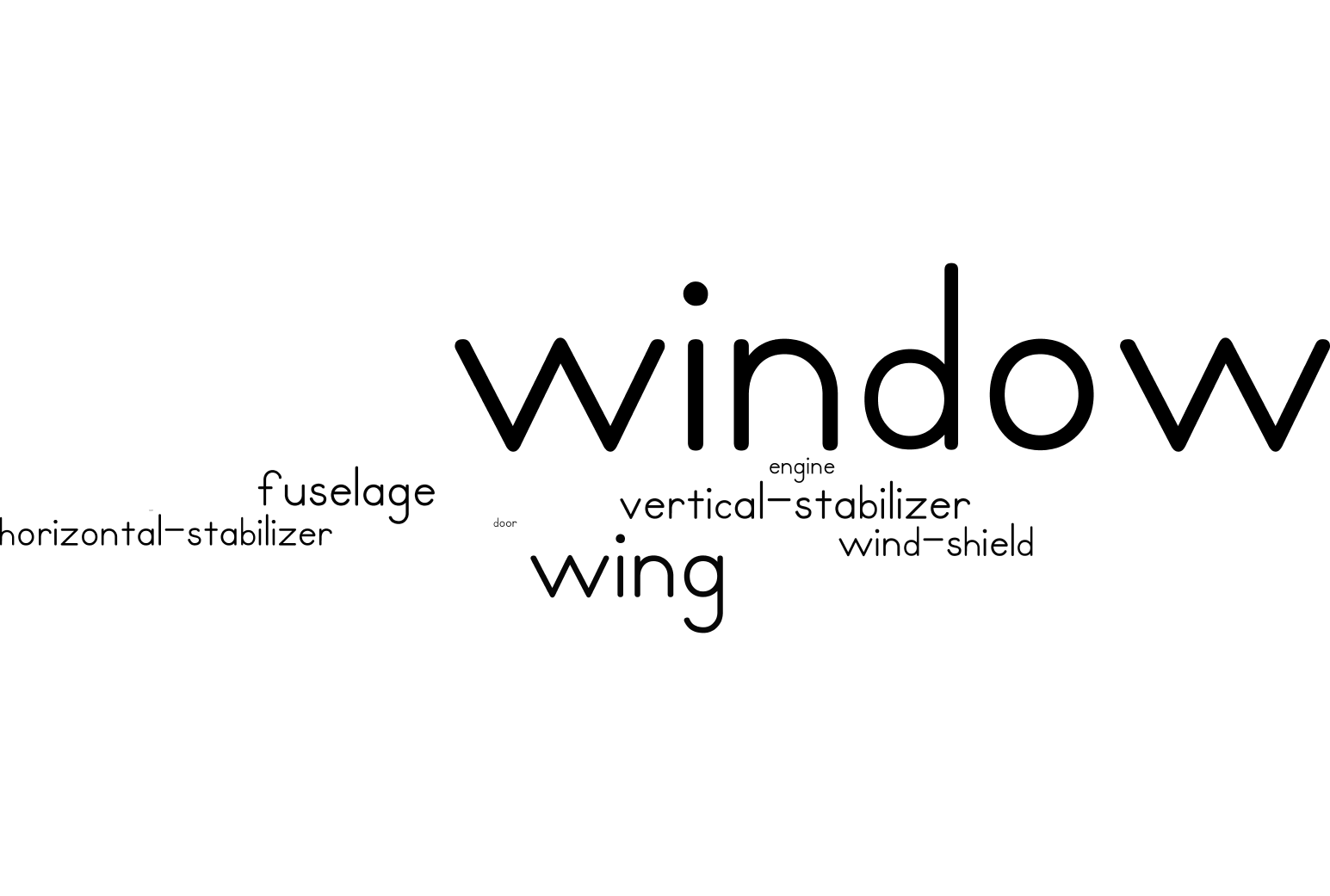}}   
						\caption*{\texttt{airplane}}
    \end{minipage}   
		\begin{minipage}{0.18\linewidth}
        \centering        
				\frame{\includegraphics[width=0.9\linewidth]{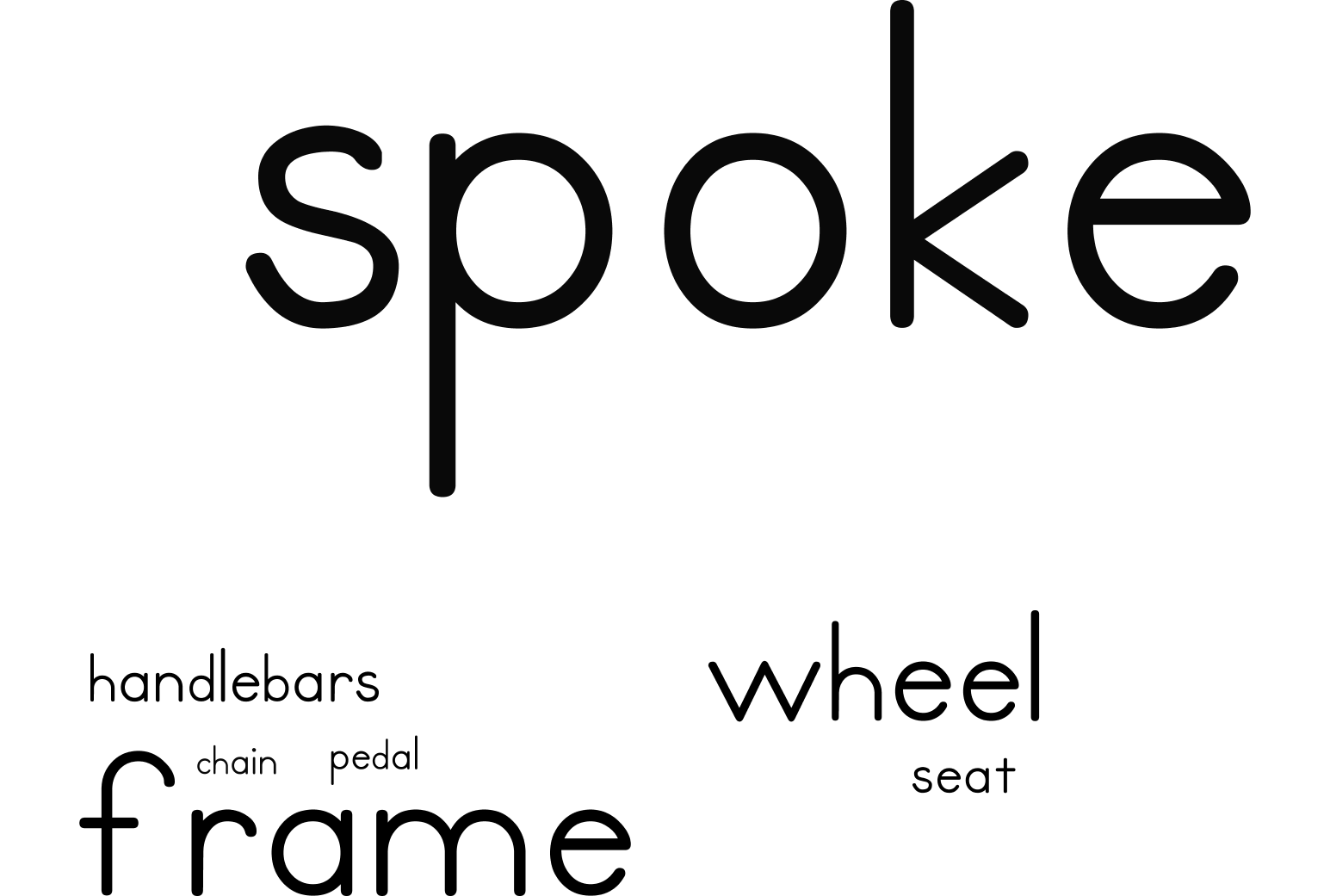}}     
					\caption*{\texttt{bicycle}}
    \end{minipage} 
		\begin{minipage}{0.18\linewidth}
        \centering        
				\frame{\includegraphics[width=0.9\linewidth]{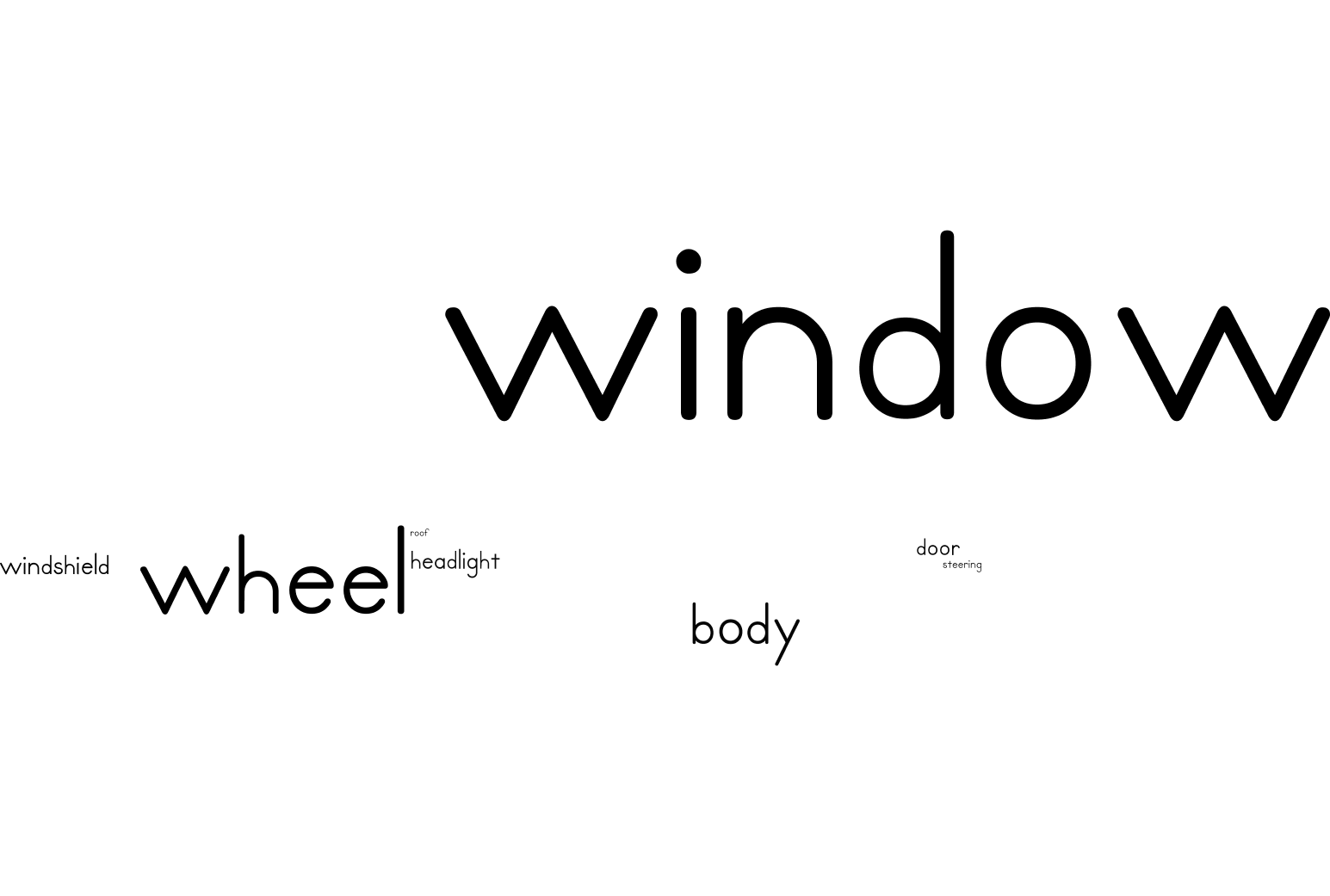}}     
					\caption*{\texttt{bus}}
    \end{minipage} 
		\begin{minipage}{0.18\linewidth}
        \centering        
				\frame{\includegraphics[width=0.9\linewidth]{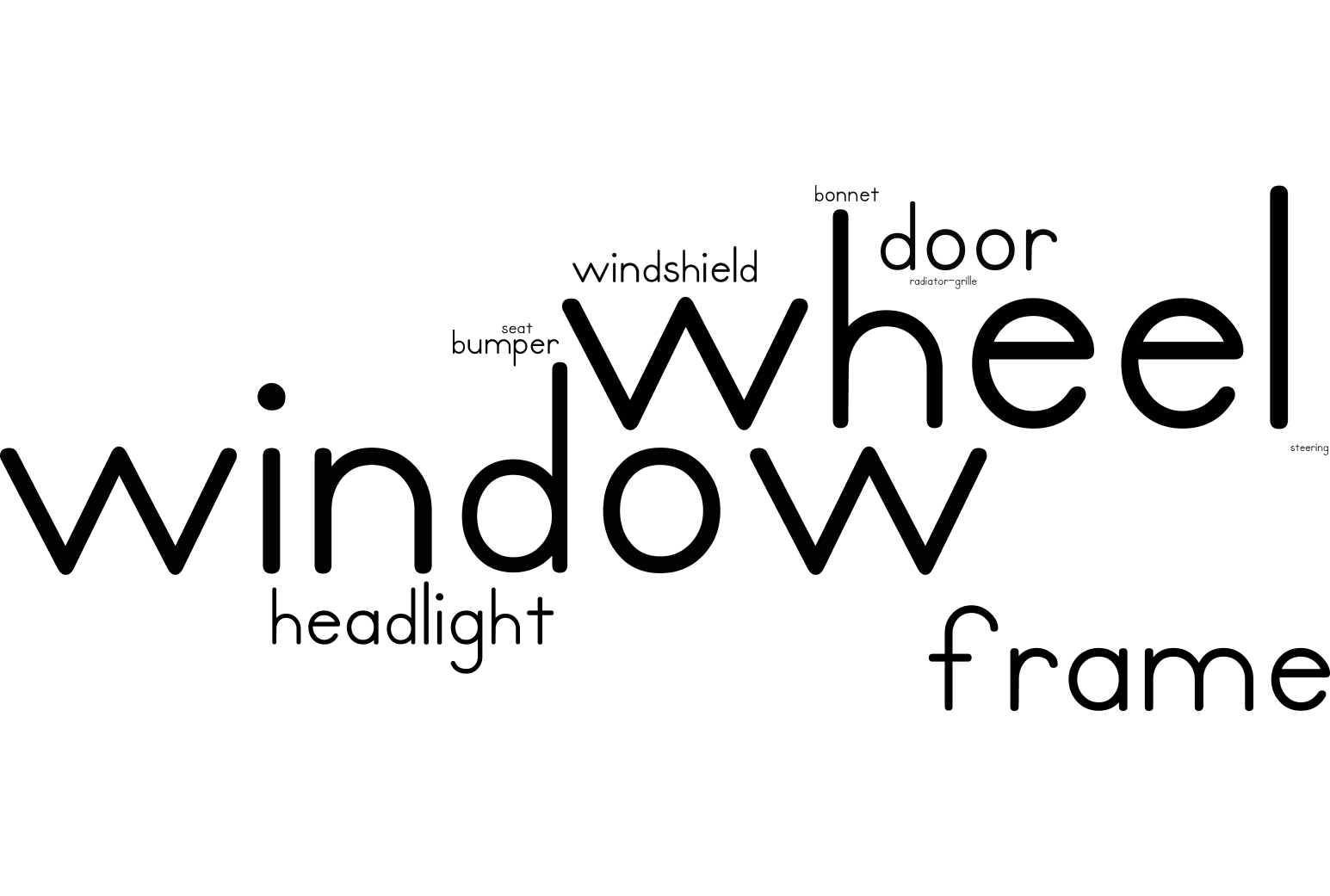} } 
						\caption*{\texttt{car}}
    \end{minipage}   
		\begin{minipage}{0.18\linewidth}
        \centering        
				\frame{\includegraphics[width=0.9\linewidth]{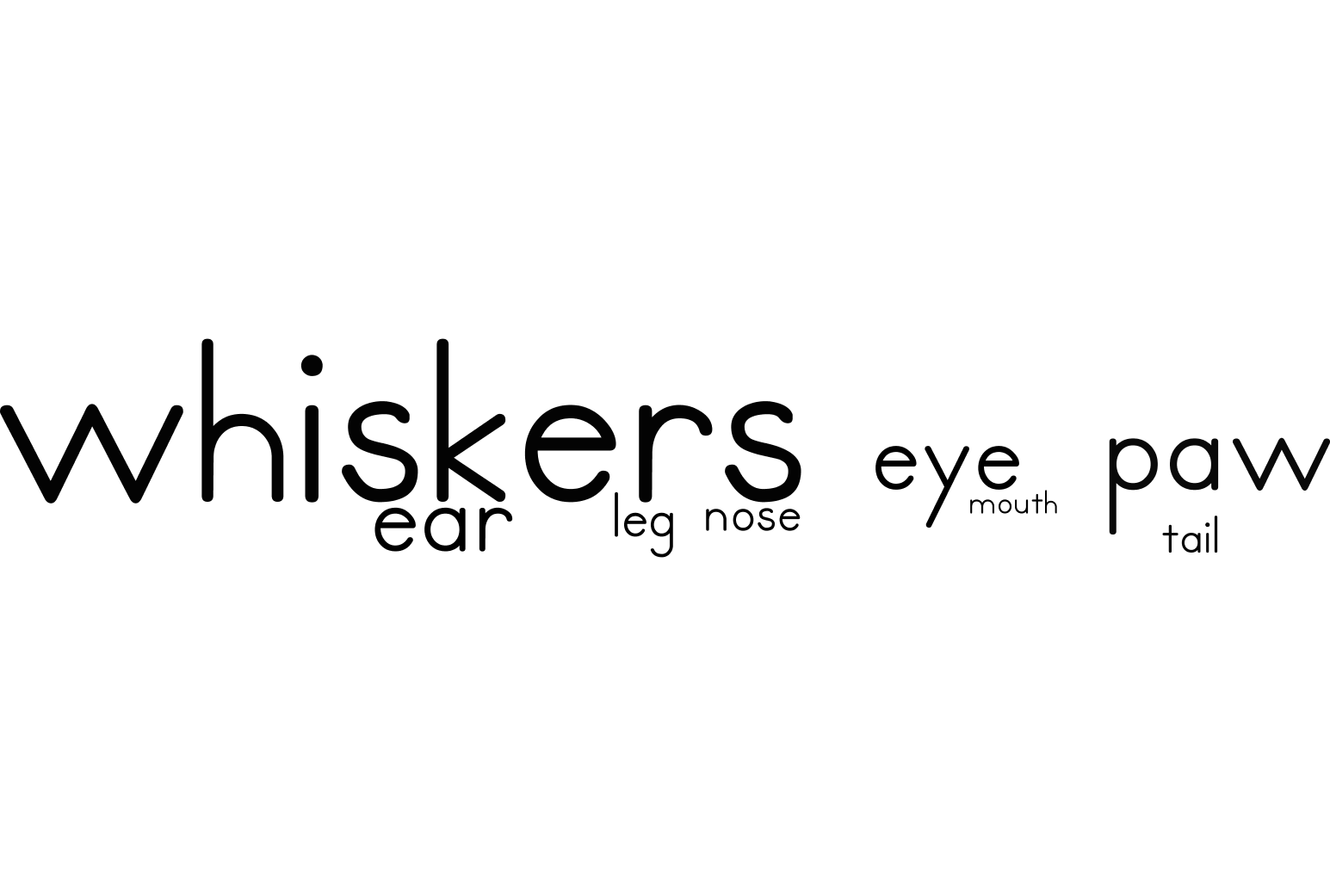} } 
						\caption*{\texttt{cat}}
    \end{minipage} 
		\vspace{5mm}
		\begin{minipage}{0.18\linewidth}
        \centering        
				\frame{\includegraphics[width=0.9\linewidth]{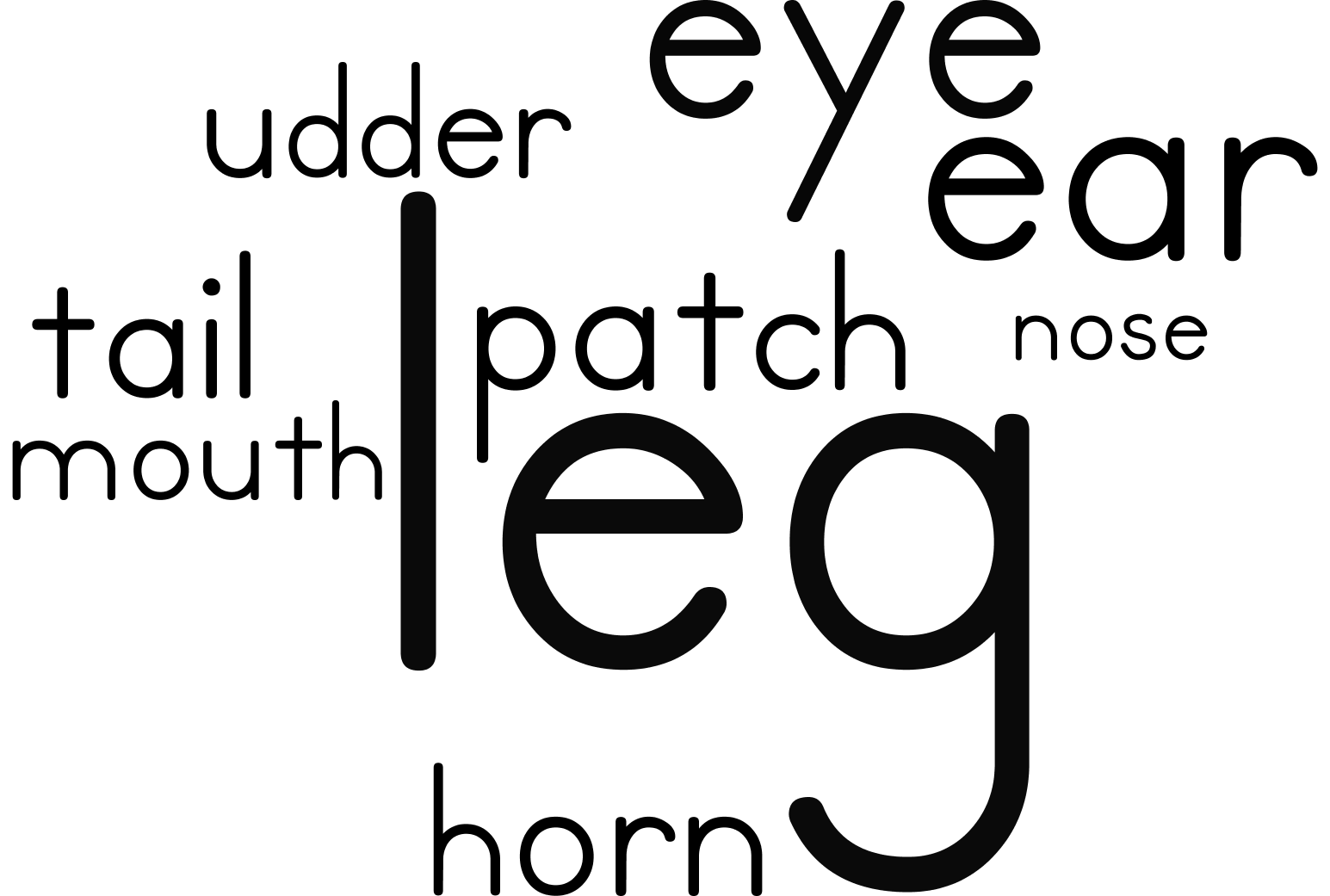}  }  
				\caption*{\texttt{cow}}
    \end{minipage}
		\begin{minipage}{0.18\linewidth}
        \centering        
				\frame{\includegraphics[width=0.9\linewidth]{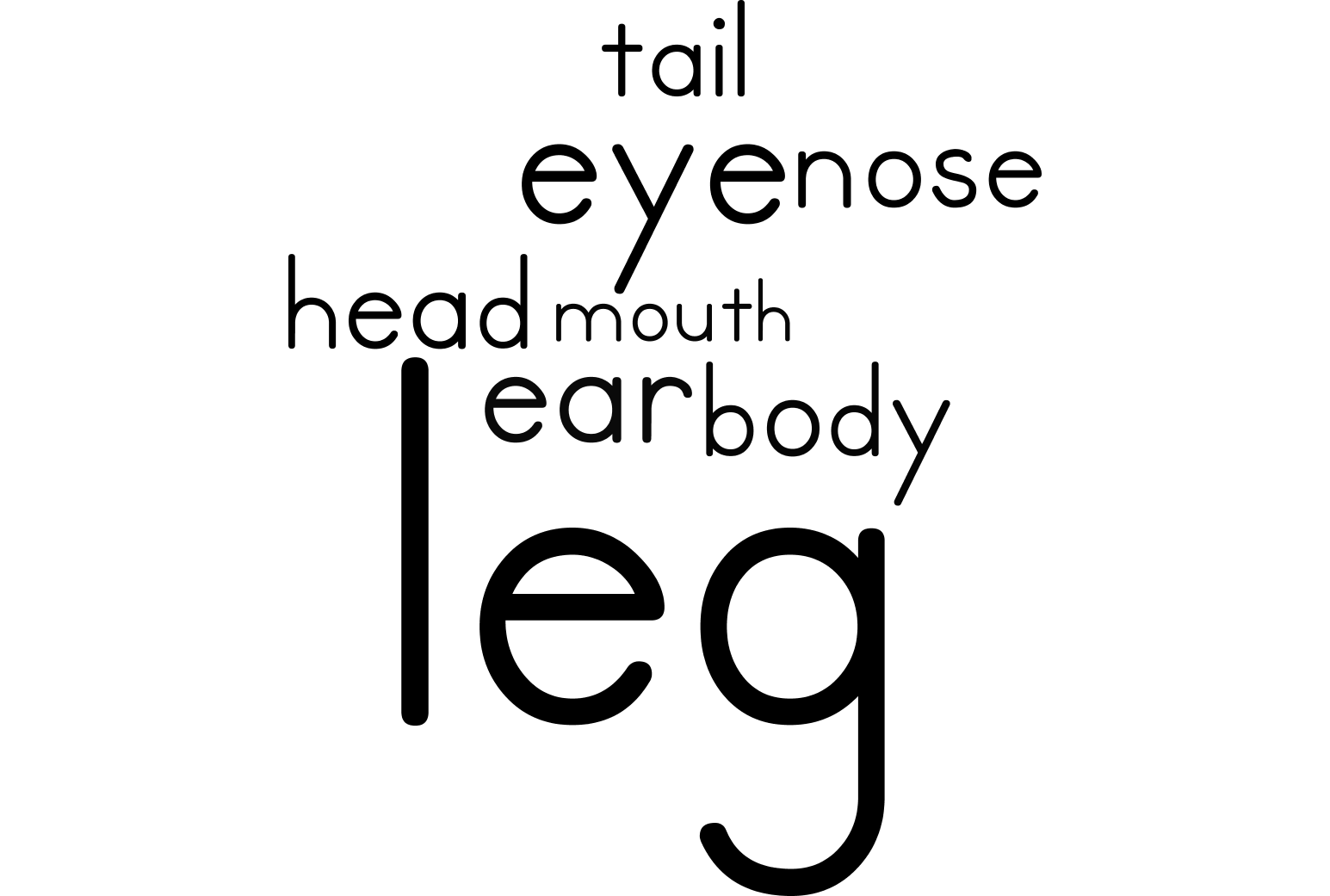}  }  
				\caption*{\texttt{dog}}
    \end{minipage} 
		\begin{minipage}{0.18\linewidth}
        \centering        
				\frame{\includegraphics[width=0.9\linewidth]{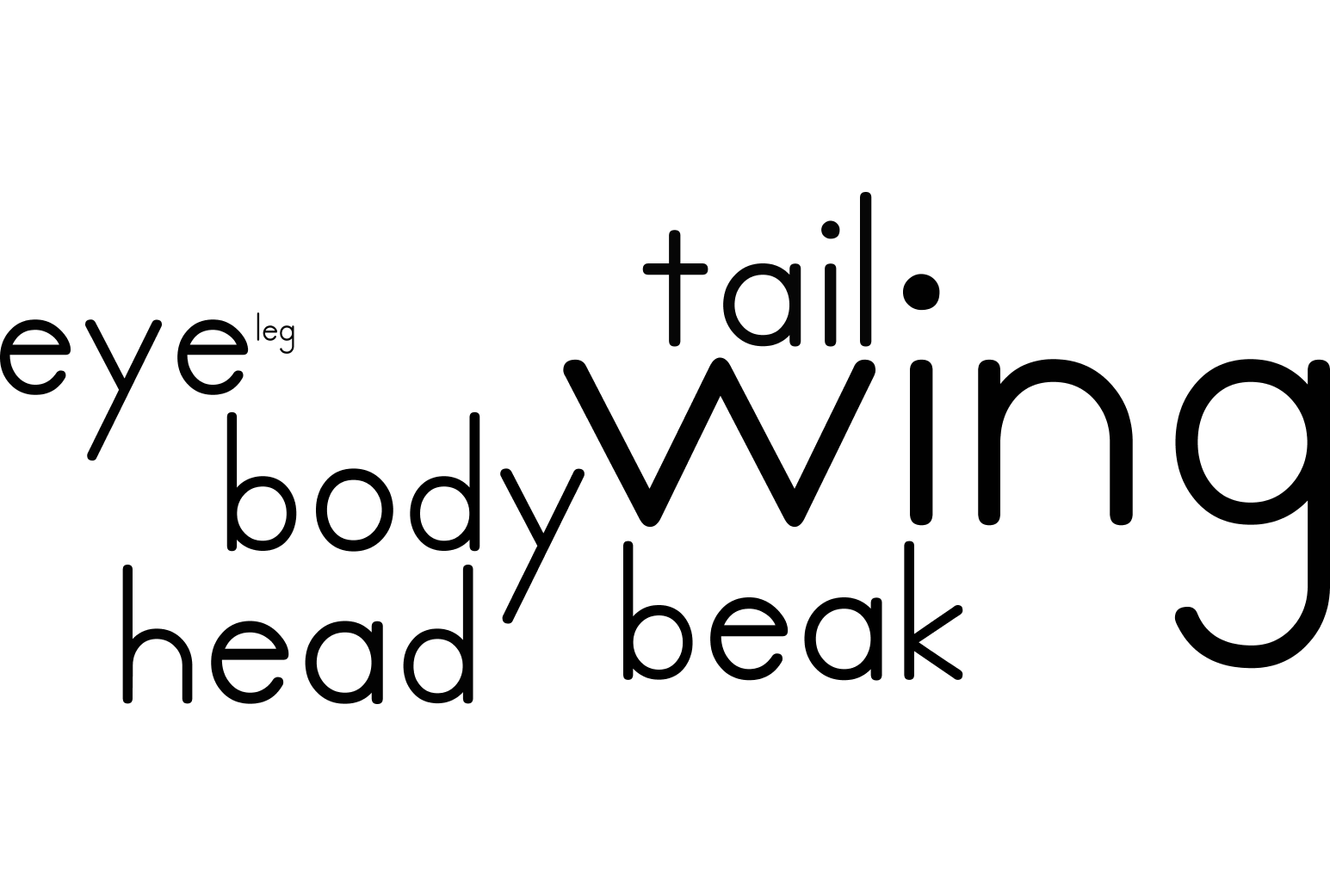}}       
				\caption*{\texttt{flying bird}}
    \end{minipage}  
		\begin{minipage}{0.18\linewidth}
        \centering        
				\frame{\includegraphics[width=0.9\linewidth]{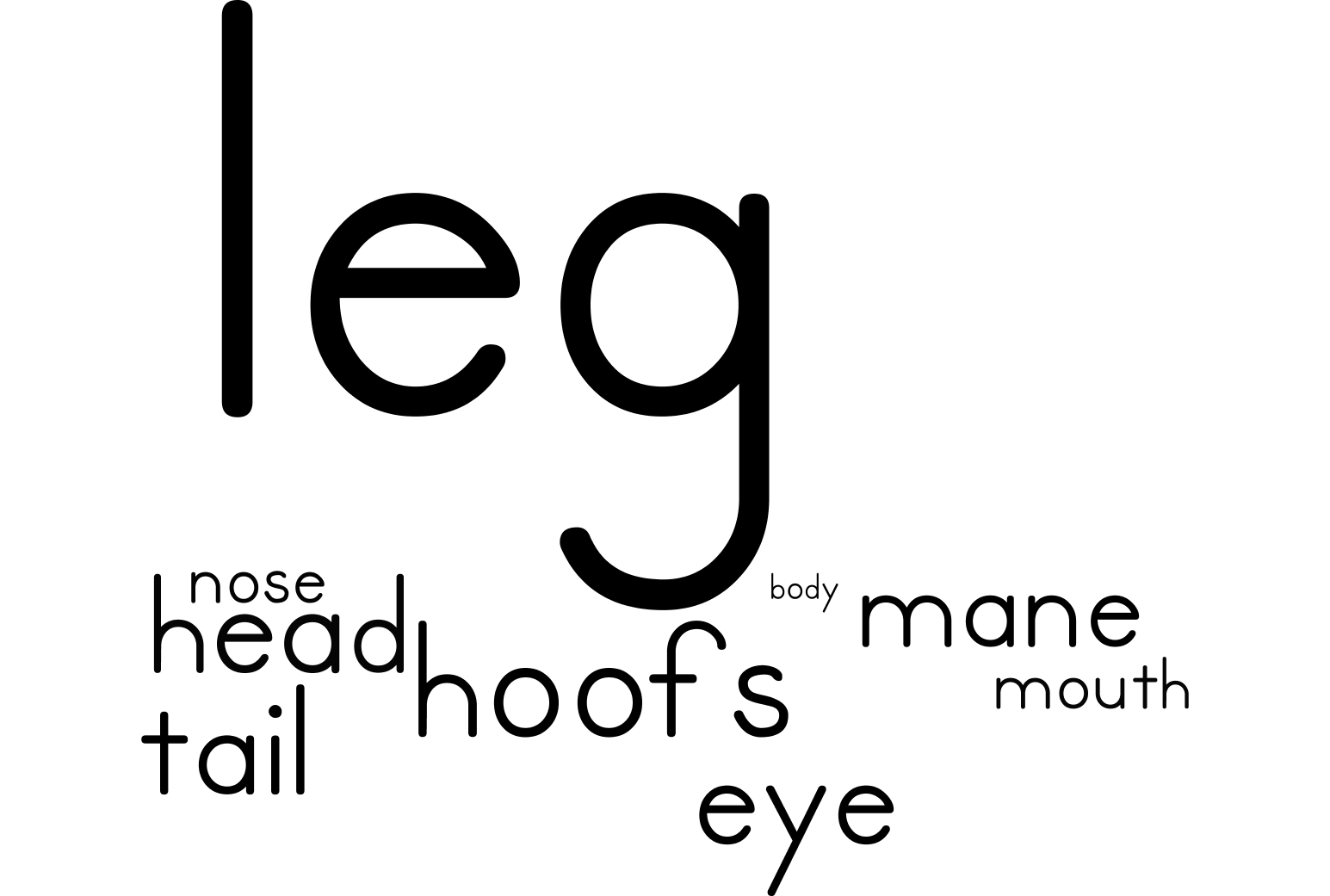}}       
				\caption*{\texttt{horse}}
    \end{minipage} 
		\begin{minipage}{0.18\linewidth}
        \centering        
				\frame{\includegraphics[width=0.9\linewidth]{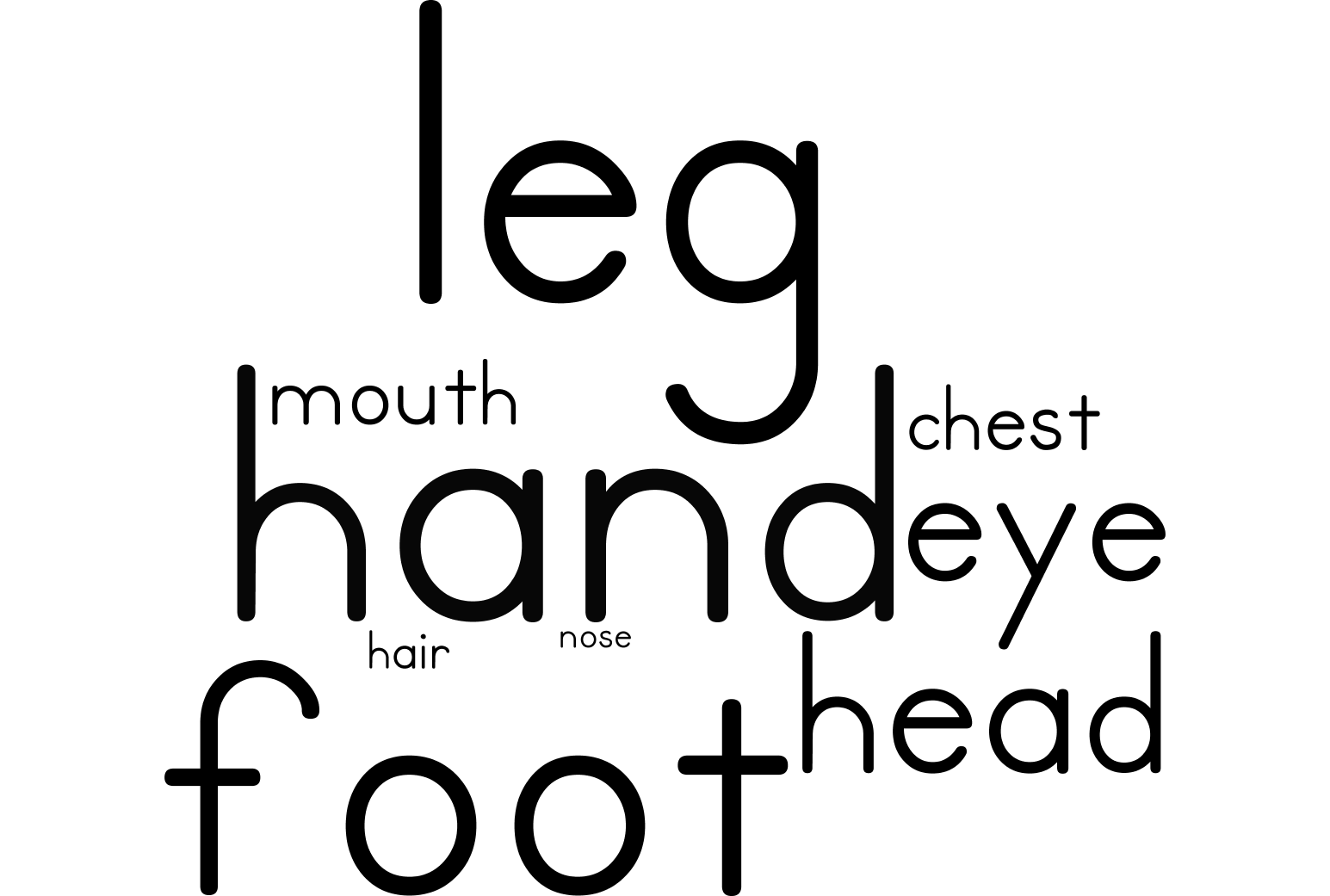}}       
				\caption*{\texttt{person walking}}
    \end{minipage} 
		\vspace{5mm}
    \centering			
		\begin{minipage}{0.18\linewidth}
        \centering        
				\frame{\includegraphics[width=0.9\linewidth]{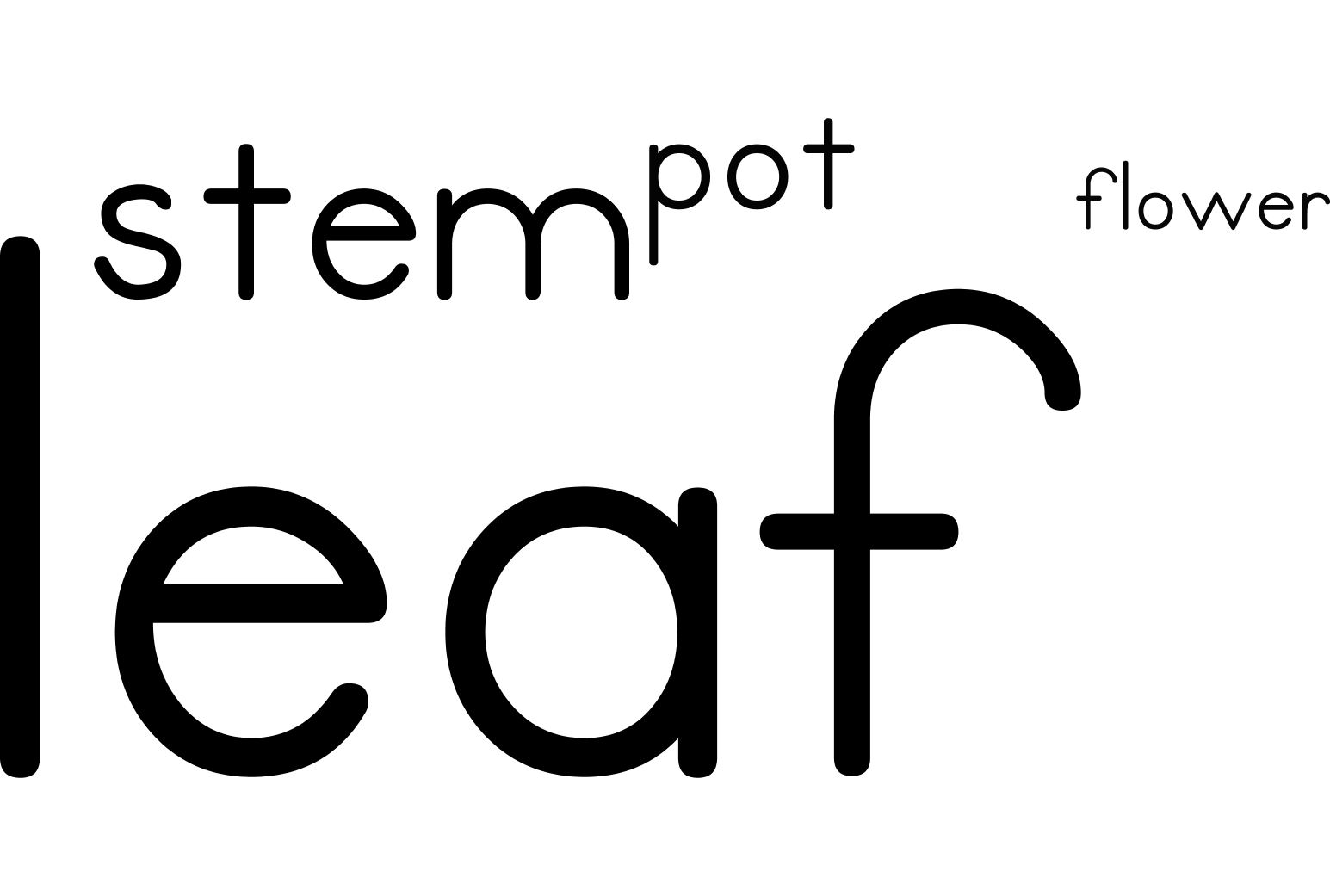}}       
				\caption*{\texttt{potted plant}}
    \end{minipage} 
		\begin{minipage}{0.18\linewidth}
        \centering        
				\frame{\includegraphics[width=0.9\linewidth]{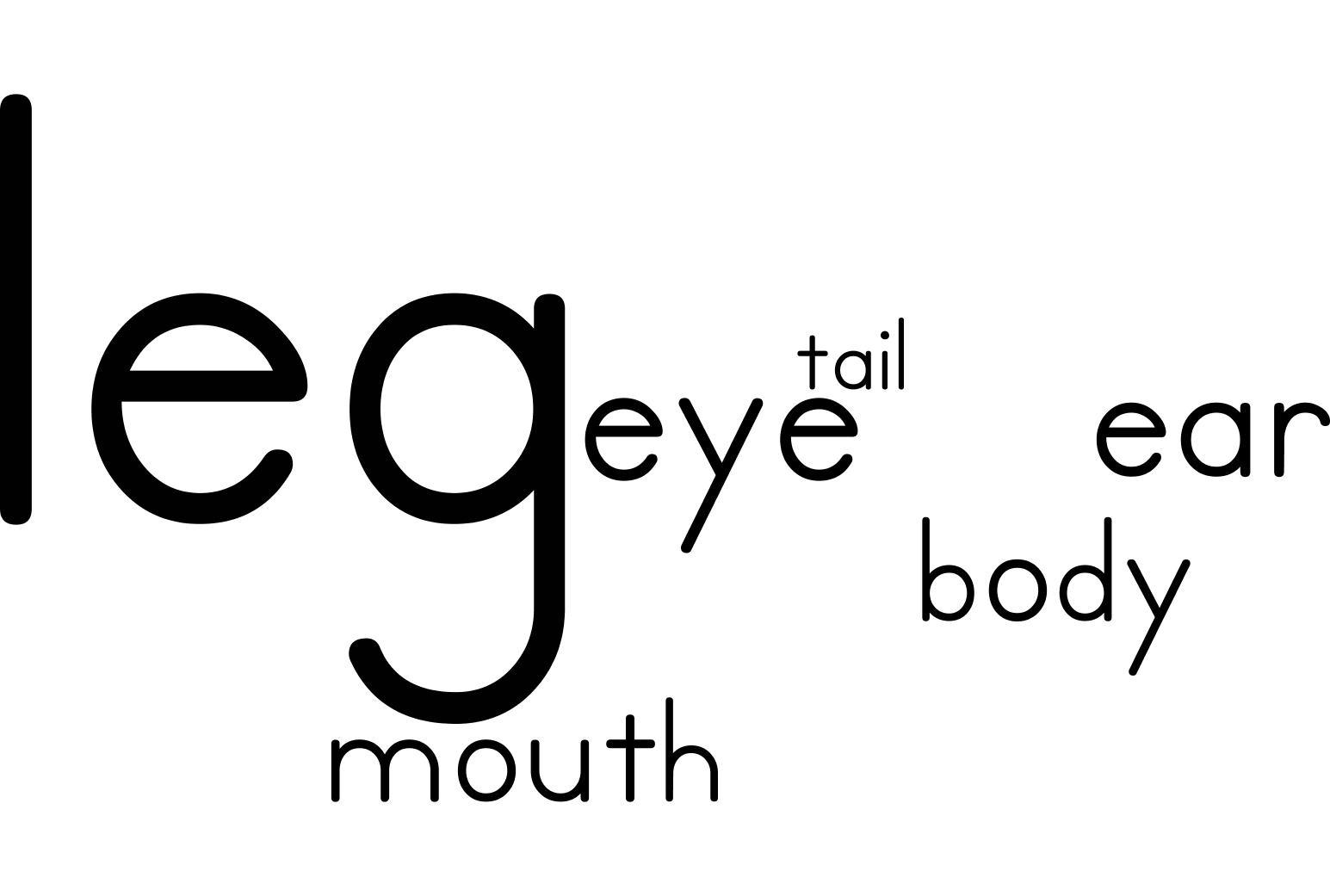}}       
				\caption*{\texttt{sheep}}
    \end{minipage} 
		\begin{minipage}{0.18\linewidth}
        \centering        
				\frame{\includegraphics[width=0.9\linewidth]{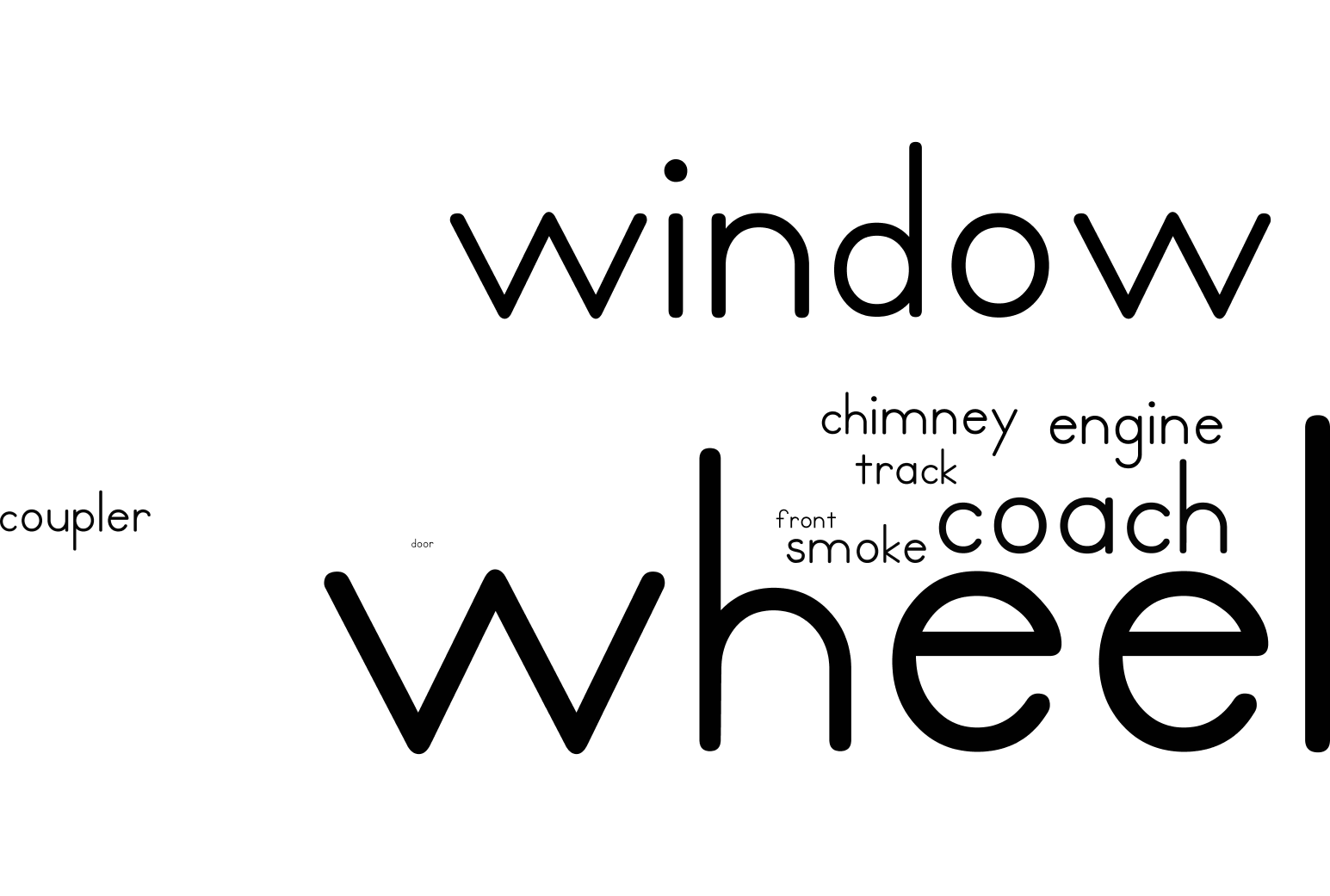}}       
				\caption*{\texttt{train}}
    \end{minipage} 
		\caption{Importance of semantic structural parts for object categories : Each image shows a word cloud of parts for epitomes of each category. The size of the part name indicates its relative importance across epitomes of the category. The above depictions are for \textsc{Temporal} stroke sequence ordering.}	
\label{fig:epitome-wordle-temporal}
\end{figure*}

\begin{figure*}[!htbp]
    \centering		
		\begin{minipage}{0.18\linewidth}
        \centering        
				\frame{\includegraphics[width=0.9\linewidth]{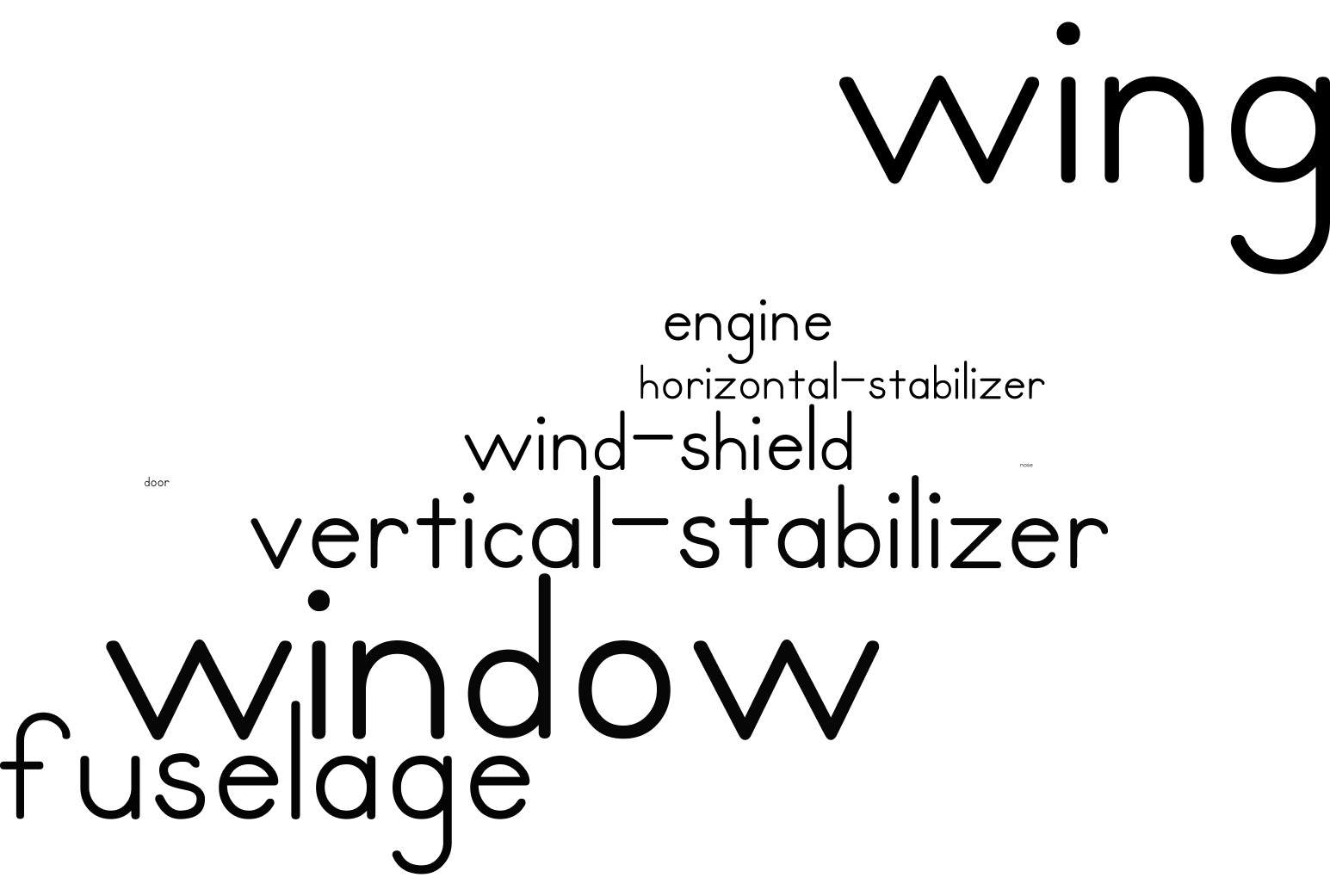}}   
						\caption*{\texttt{airplane}}
    \end{minipage}   
		\begin{minipage}{0.18\linewidth}
        \centering        
				\frame{\includegraphics[width=0.9\linewidth]{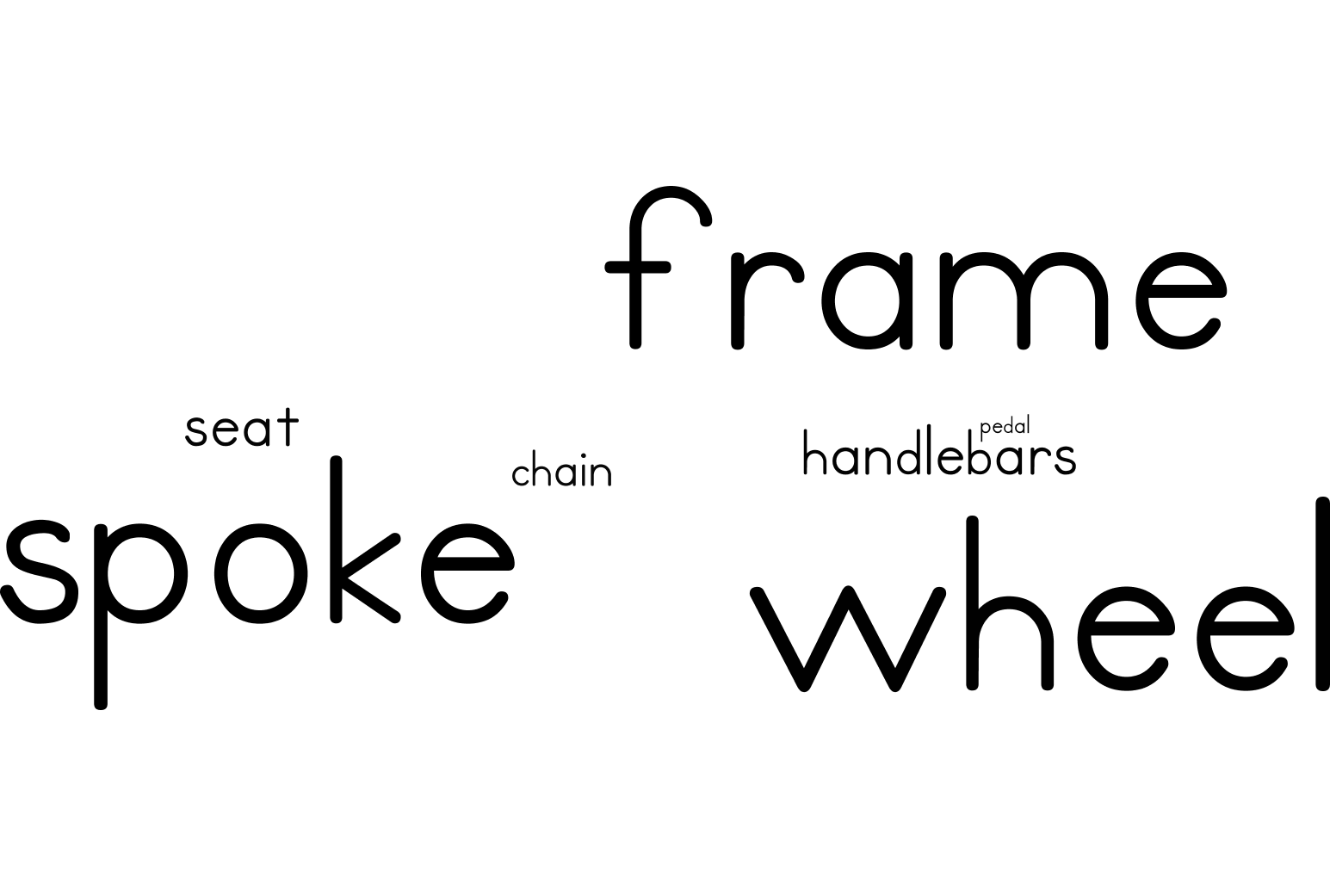}}     
					\caption*{\texttt{bicycle}}
    \end{minipage} 
		\begin{minipage}{0.18\linewidth}
        \centering        
				\frame{\includegraphics[width=0.9\linewidth]{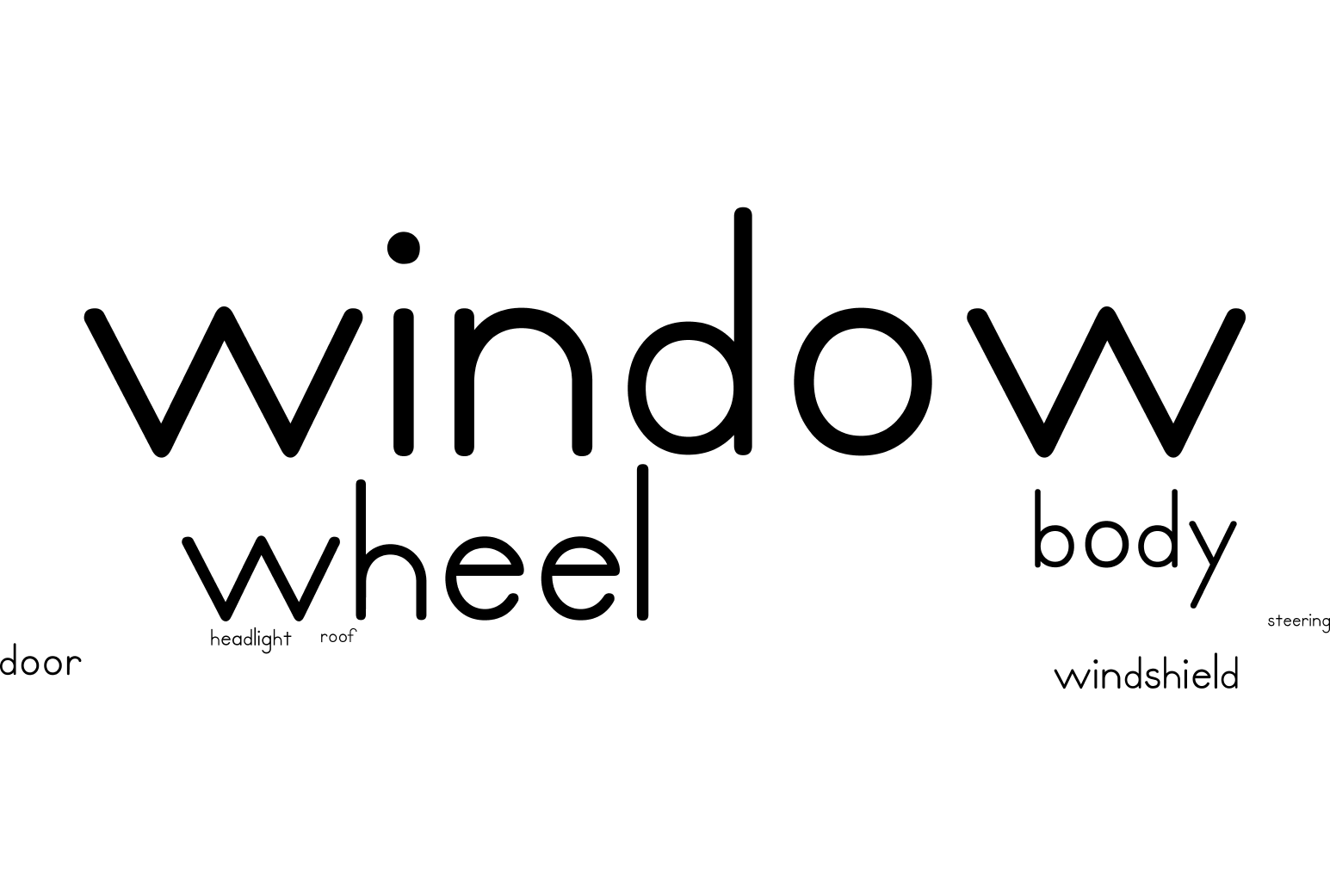}}     
					\caption*{\texttt{bus}}
    \end{minipage} 
		\begin{minipage}{0.18\linewidth}
        \centering        
				\frame{\includegraphics[width=0.9\linewidth]{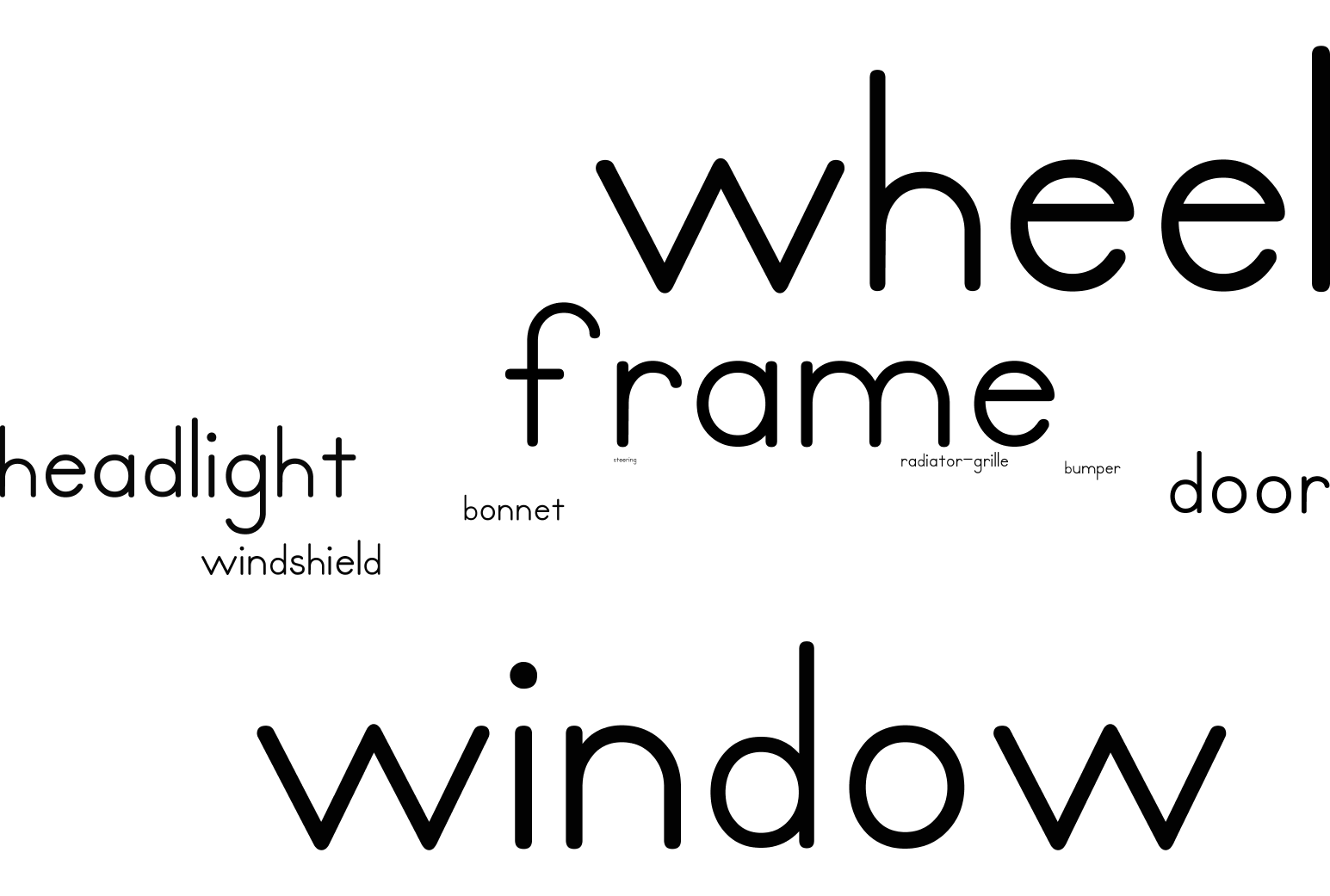} } 
						\caption*{\texttt{car}}
    \end{minipage}   
		\begin{minipage}{0.18\linewidth}
        \centering        
				\frame{\includegraphics[width=0.9\linewidth]{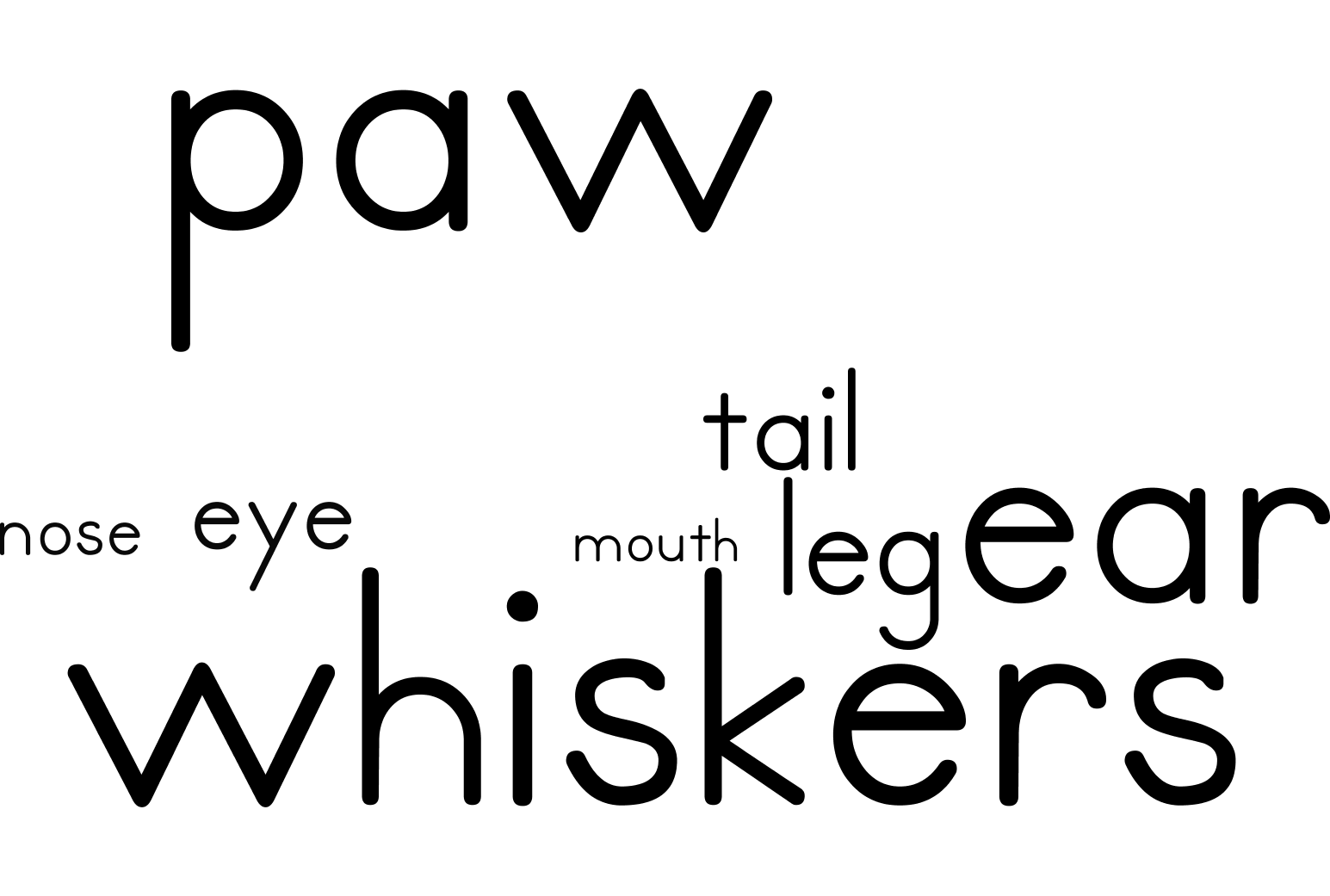} } 
						\caption*{\texttt{cat}}
    \end{minipage} 
		\begin{minipage}{0.18\linewidth}
        \centering        
				\frame{\includegraphics[width=0.9\linewidth]{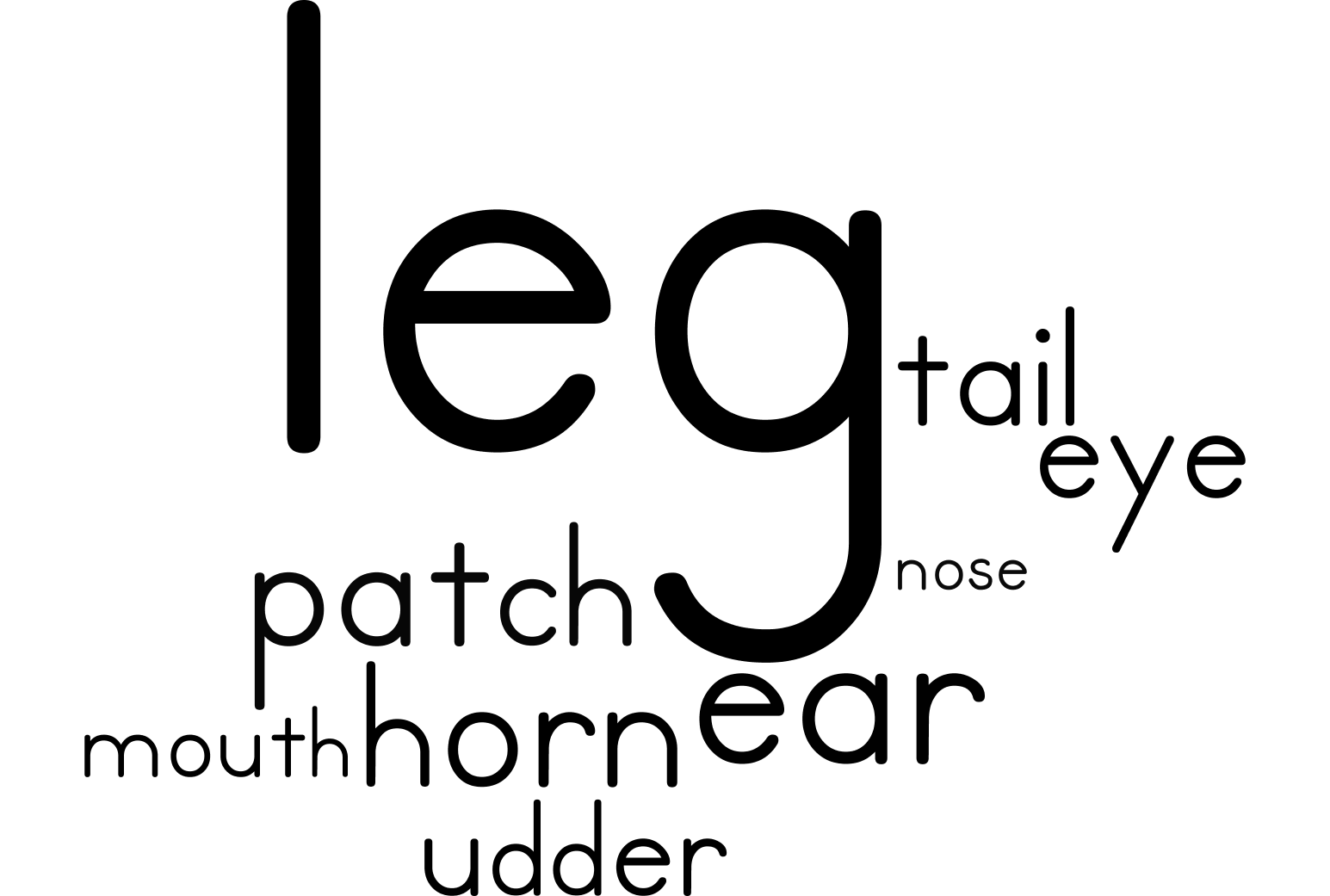}  }  
				\caption*{\texttt{cow}}
    \end{minipage}
		\begin{minipage}{0.18\linewidth}
        \centering        
				\frame{\includegraphics[width=0.9\linewidth]{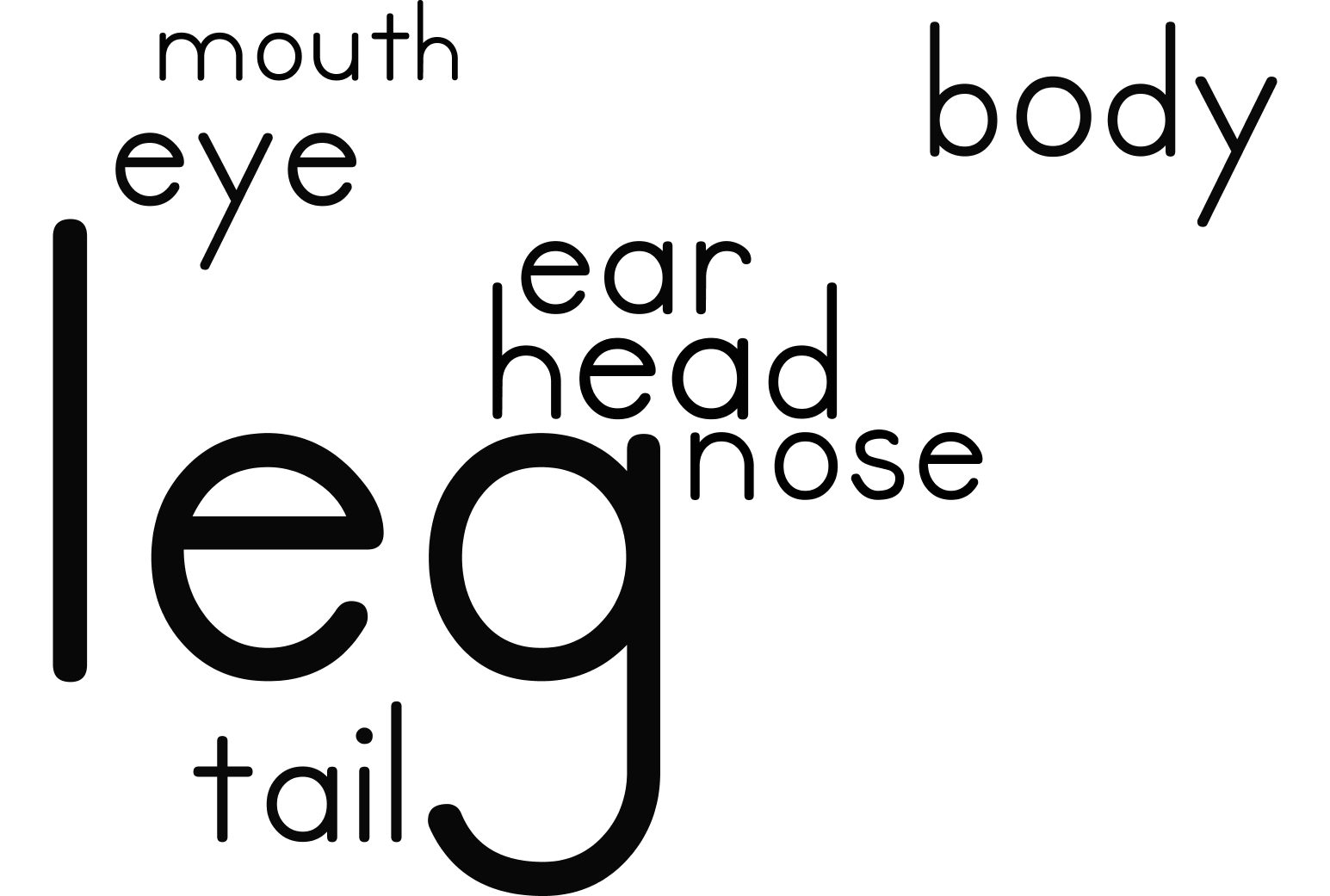}  }  
				\caption*{\texttt{dog}}
    \end{minipage} 
		\begin{minipage}{0.18\linewidth}
        \centering        
				\frame{\includegraphics[width=0.9\linewidth]{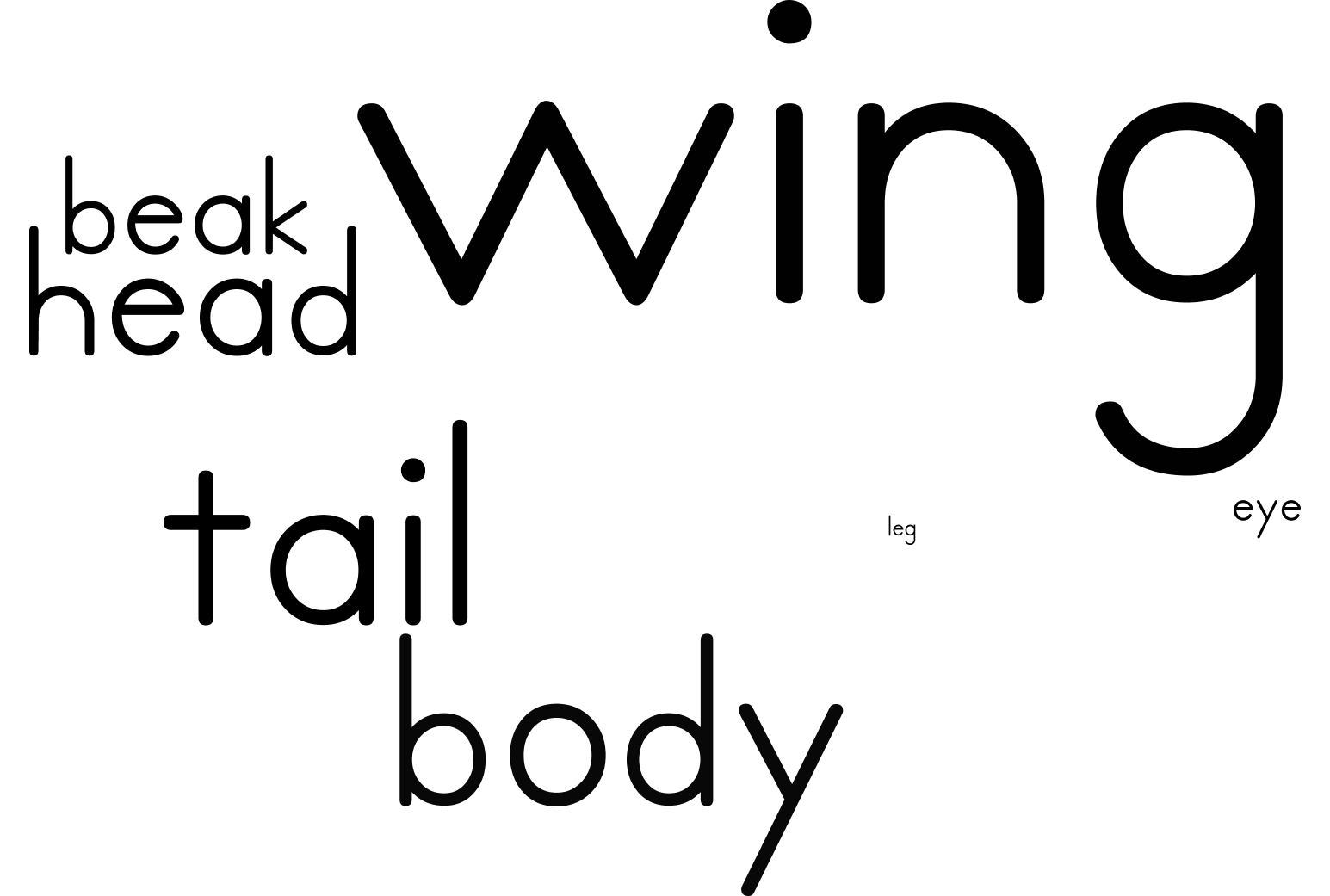}}       
				\caption*{\texttt{flying bird}}
    \end{minipage}  
		\begin{minipage}{0.18\linewidth}
        \centering        
				\frame{\includegraphics[width=0.9\linewidth]{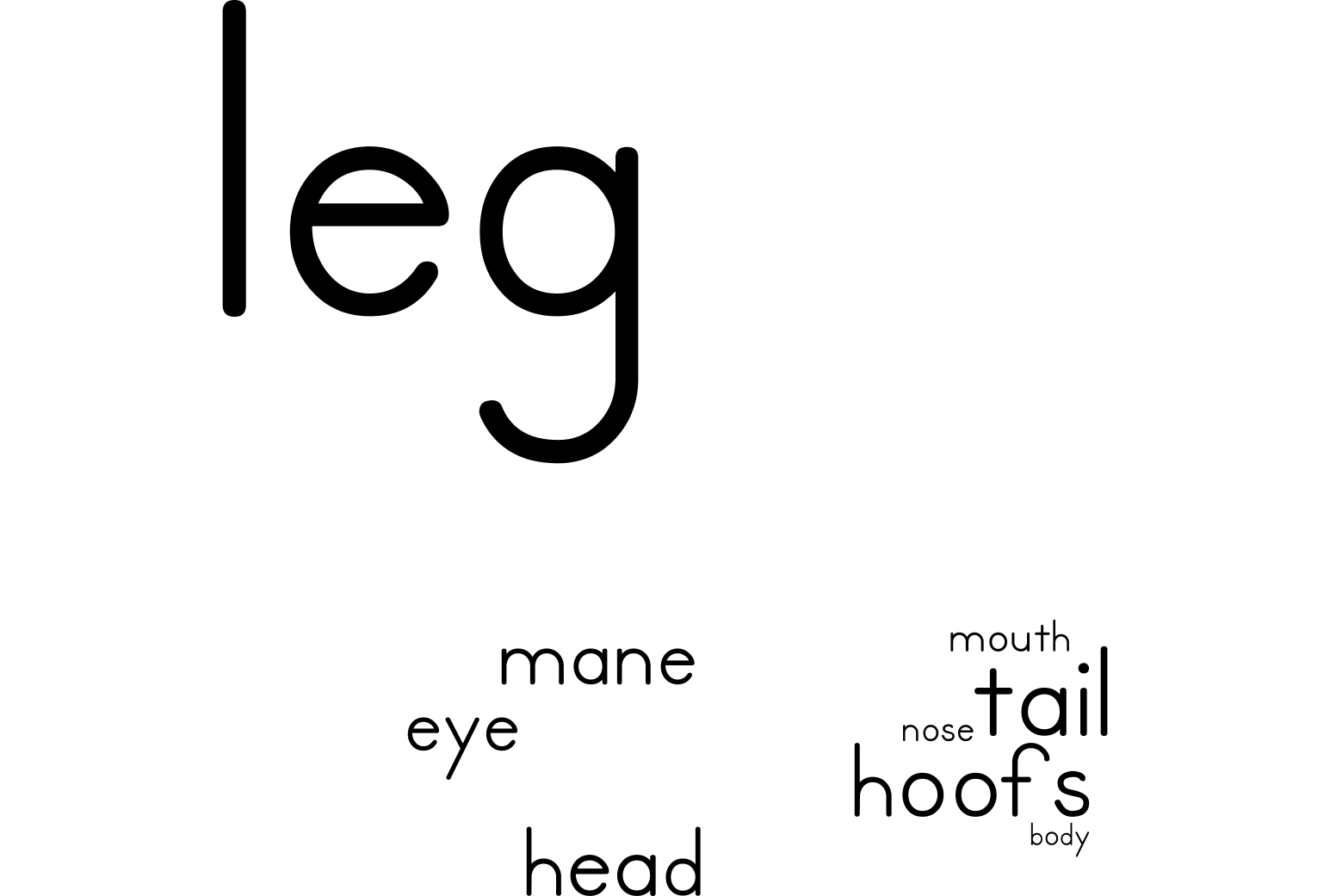}}       
				\caption*{\texttt{horse}}
    \end{minipage} 
		\begin{minipage}{0.18\linewidth}
        \centering        
				\frame{\includegraphics[width=0.9\linewidth]{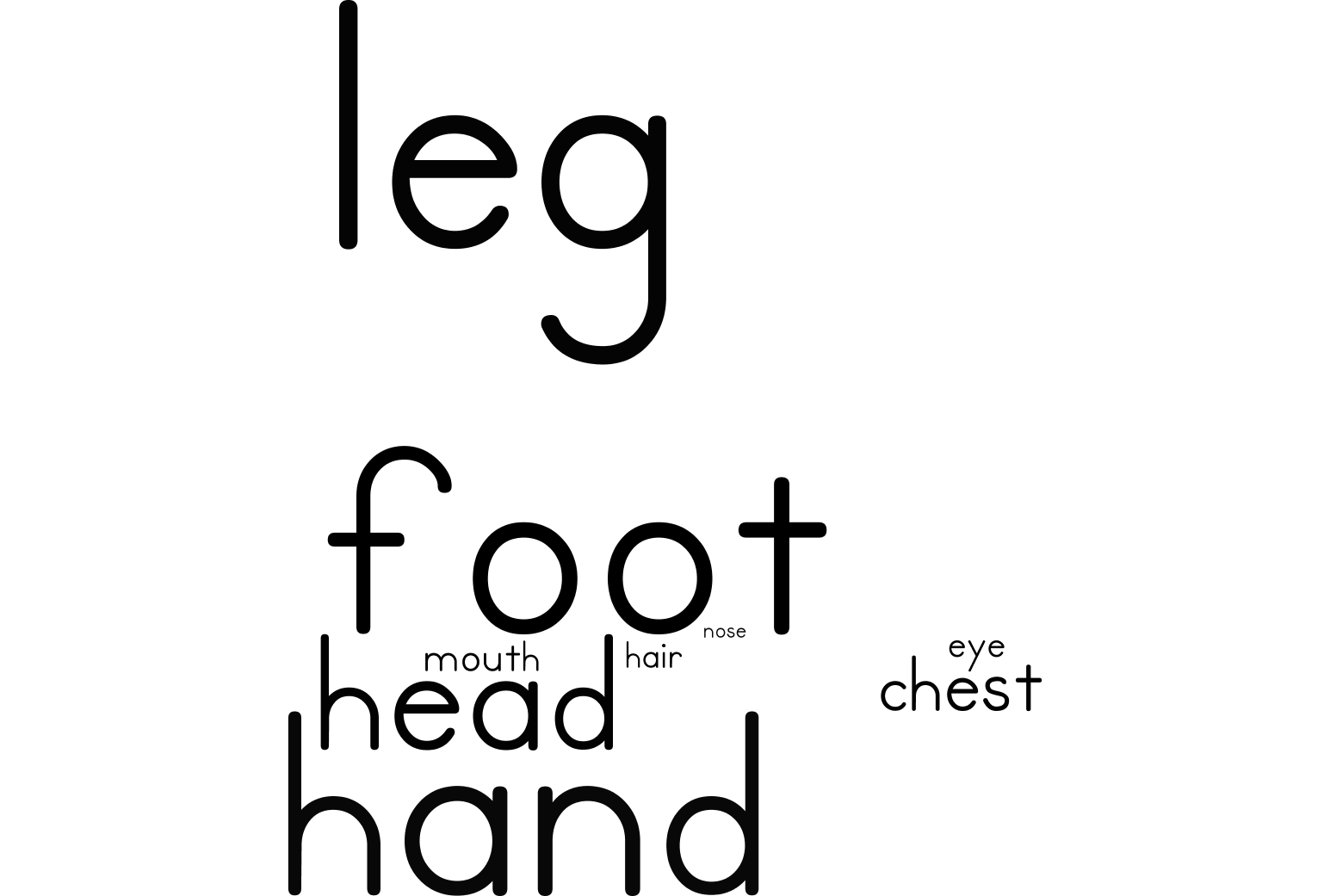}}       
				\caption*{\texttt{person walking}}
    \end{minipage} 
    \centering			
		\begin{minipage}{0.18\linewidth}
        \centering        
				\frame{\includegraphics[width=0.9\linewidth]{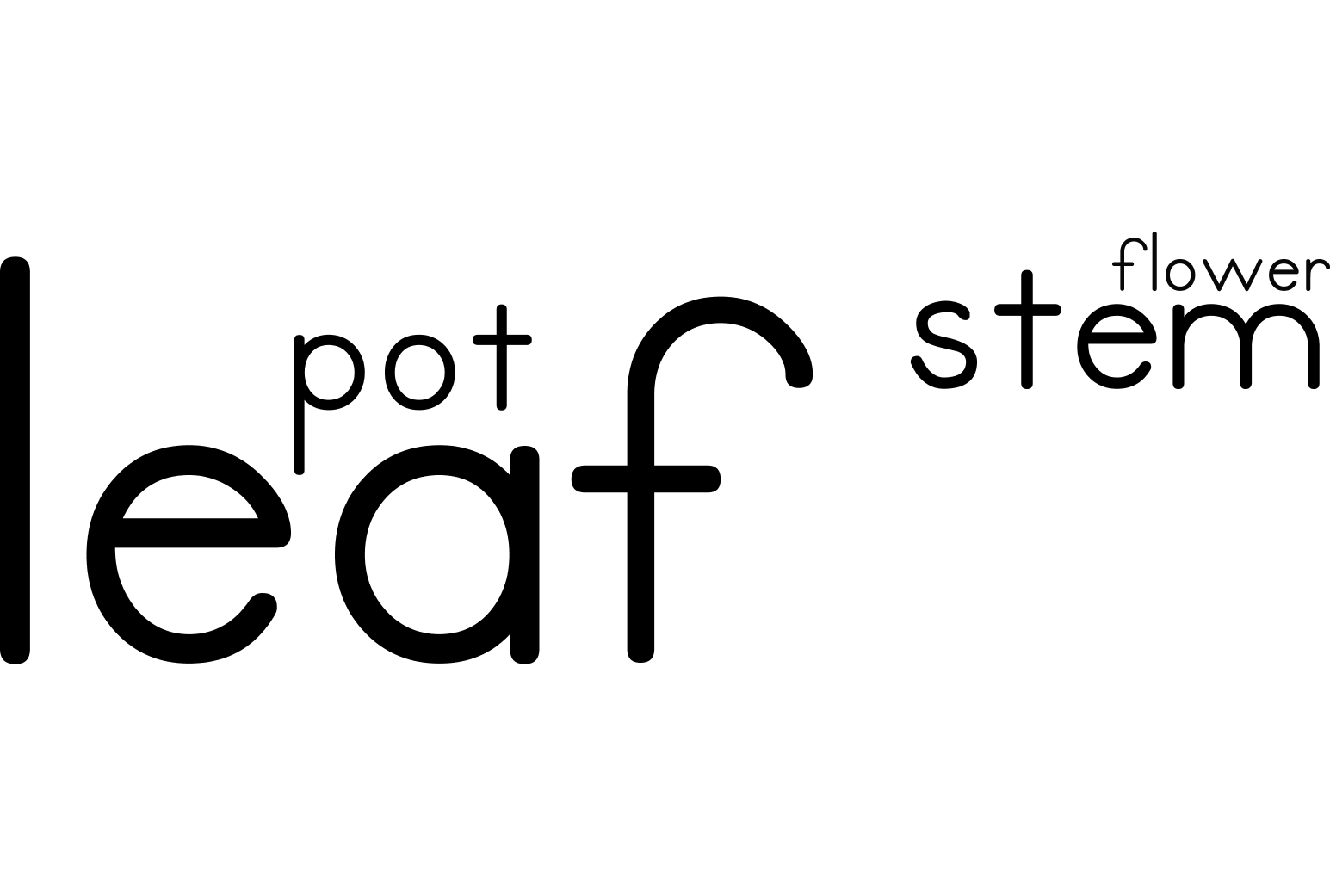}}       
				\caption*{\texttt{potted plant}}
    \end{minipage} 
		\begin{minipage}{0.18\linewidth}
        \centering        
				\frame{\includegraphics[width=0.9\linewidth]{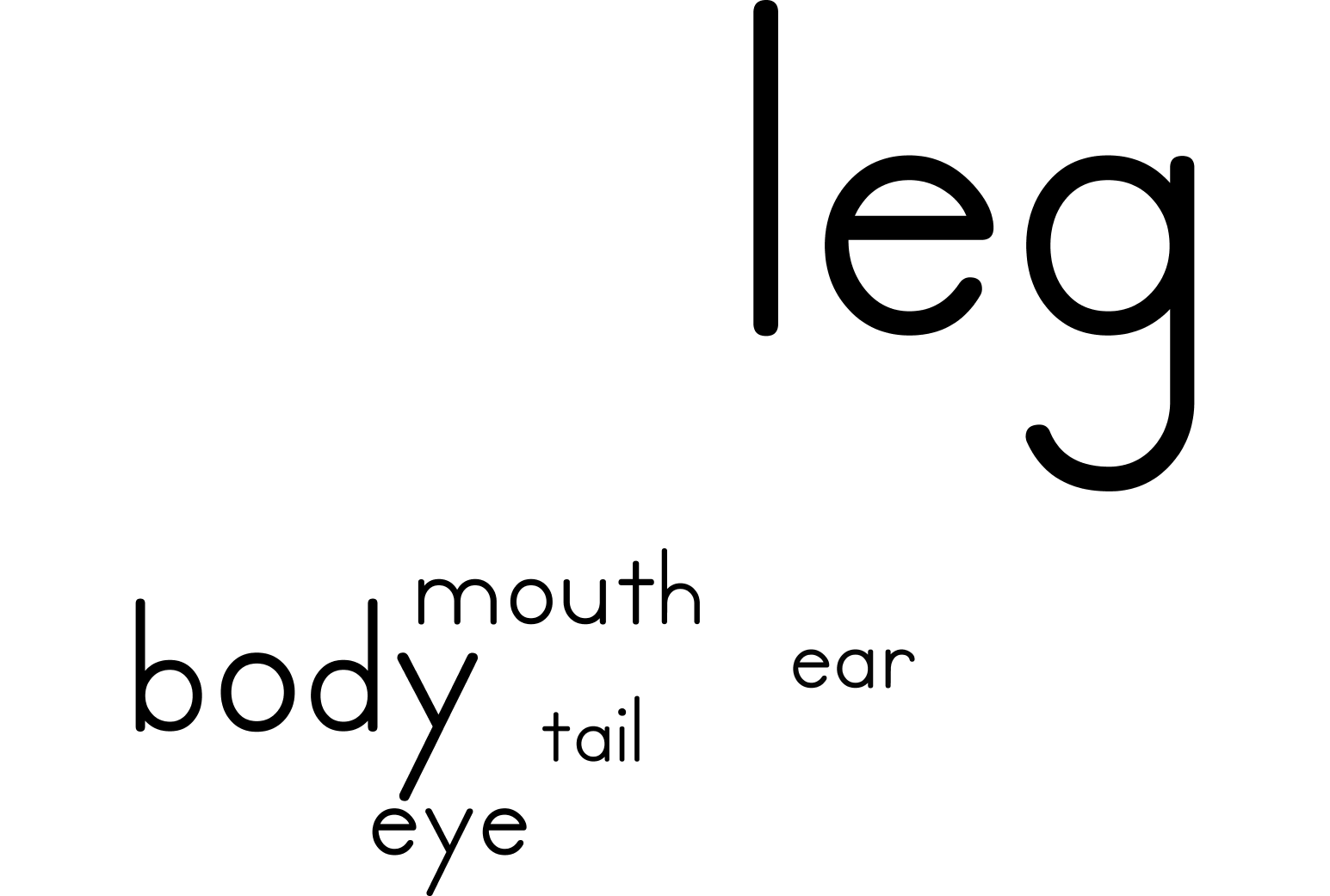}}       
				\caption*{\texttt{sheep}}
    \end{minipage} 
		\begin{minipage}{0.18\linewidth}
        \centering        
				\frame{\includegraphics[width=0.9\linewidth]{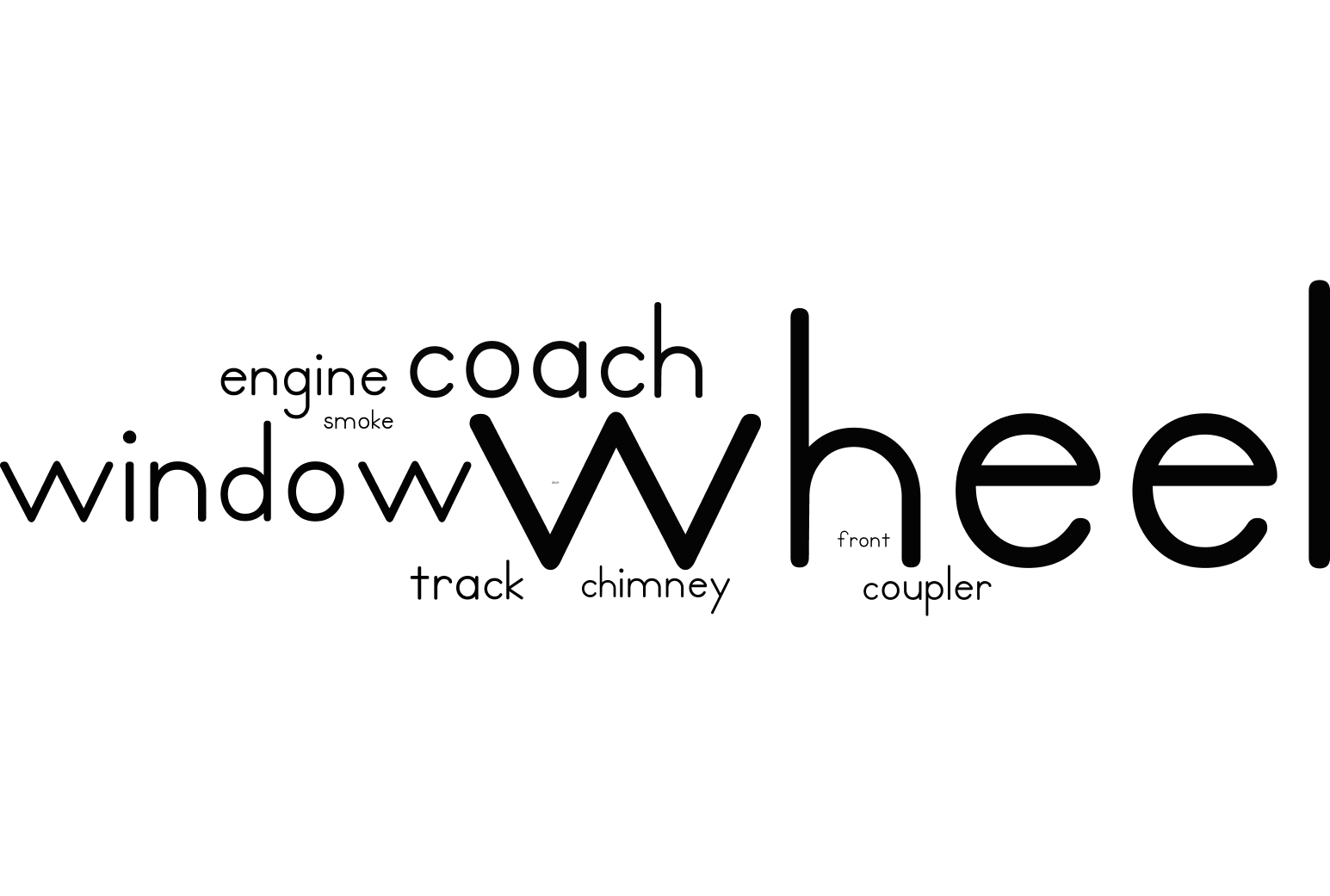}}       
				\caption*{\texttt{train}}
    \end{minipage} 
		\caption{Importance of semantic structural parts for object categories : Each image shows a word cloud of parts for epitomes of each category. The size of the part name indicates its relative importance across epitomes of the category. The above depictions are for \textsc{Alternate} stroke sequence ordering.}	
\label{fig:epitome-wordle-alternate}
\end{figure*}

\subsection{Obtaining `fine-grained' part-importance weights}
\label{sec:finegrained}

The process of annotating a part typically results in a 2-D closed contour. Points on the contour tend to be in close proximity with the boundary pixels of the object. We exploit this observation to obtain a `fine-grained' part importance factor for each part enclosed by a candidate contour.

Let $B^{(j)} = \{ b^{(j)}_i = (x_i^{(j)},y_i^{(j)})\}_{i=1}^{n_b}, j = 1 \ldots n_a$ be the sets of points comprising user-annotated boundary contours. Examples of such contours can be in the annotation image $\mathbb{A}$ in Figure \ref{fig:overview} (region \circlednumber{3}). For a given `part contour set' $B^{(j)}$, let $S^{(j)} = \{s^{(j)}_k  =  (x_k^{(j)},y_k^{(j)})\}_{k=1}^{n_s}, j = 1 \ldots n_a$ be the `full-sketch point' set of 2-D locations of stroke pixels enclosed by the part contour in the full sketch $\mathbb{S}$.  Similarly, let $E^{(j)} = \{e^{(j)}_l =  (x_l^{(j)},y_l^{(j)})\}_{l=1}^{n_e}, j = 1 \ldots n_a$ be the `epitome point' set of 2-D locations of stroke pixels enclosed by the part contour in epitome $\mathbb{E}$. 

For each member in the `full-sketch point' set, $s^{(j)}_{k} = (x_k^{(j)},y_k^{(j)}) \in S^{(j)}$, we find the closest-by-distance matching point $b^{(j)}_{i} = (x_i^{(j)},y_i^{(j)})$ from $B^{(j)}$. i.e. $b^{(j)}_{i} = \argmin_x  d(s^{(j)}_{k},b^{(j)}_{x})$. We retain only those matches $s^{(j)}_{k}, b^{(j)}_{i}$ whose distance is less than a threshold. Intuitively, this procedure enables us to identify stroke pixels of the full sketch which ``hug'' the candidate contour's boundary. Let $n_{valid\_full}^{(j)}$ be the number of such stroke pixels $s^{(j)}_{k}$. Valid matches found using the above procedure are shown in red for a candidate part boundary in Figure \ref{fig:overview} -- refer to region \circlednumber{4}. A similar procedure gives us $n_{valid\_epi}^{(j)}$ -- the number of stroke pixels in the epitome which ``hug" the candidate contour's boundary (See region \circlednumber{5} in Figure ~\ref{fig:overview}). The ratio $w_{part}^{(j)} = \frac{n_{valid\_epi}^{(j)}}{n_{valid\_full}^{(j)}}$ provides a fine-grained importance for the part enclosed by the boundary -- the higher the value of $w_{part}^{(j)}$, the more the number of pixels from both full sketch and the epitome which commonly ``hug" the annotated part boundary. The part-importance is depicted in Figure \ref{fig:overview} by the height of cylinders adjoining numbered regions \circlednumber{7},\circlednumber{8}. As we can see, the `wheel' is more prominently present in the epitome compared to the `bicycle frame' thereby according the former a larger part-importance ($w_{part}^{(j)}$) value.

\subsection{Obtaining category-wise part weights}

The above procedure of obtaining fine-grained importance weight $w_{part}^{(j)}$ is repeated for each part (indexed by $j = 1,2\ldots n_a$) enclosed by a candidate contour boundary $B^{(j)}$ by utilizing pixel location sets $S^{(j)}$ and $E^{(j)}$. These weights are combined along with the coarse importance weights $p_{im}$ obtained in Section \ref{sec:candidate} to obtain the image-level part-wise importance weights $f_i,i=1,2,\ldots M$ (region labeled \circlednumber{6} in Figure \ref{fig:overview}). The part-wise aggregation of these weights over all the `full sketch'-epitome pairs of a category results is normalized to obtain a probability distribution of part-level importance. This distribution can then be visualized as a semantic-part word cloud (see Figure \ref{fig:epitomestowordle}) to determine the ``signature" structural elements (semantic-parts) which persist in the epitomes of a category.

The pseudo-code for the procedure described in this section can be viewed in Appendix \ref{sec:pseudocode}. In the pseudo-code, portions highlighted in red indicate procedures whose details are provided separately.

\section{Analyzing semantic-part word clouds}
\label{sec:analysis}

Before we proceed, it is important to point out that the \textit{category-epitome} of a full sketch implicitly depends upon the order in which individual strokes of the full sketch are considered. For example, the epitome can be constructed by considering the  temporal order in which the sketches were originally added. Yet another epitome can be constructed if we consider strokes in decreasing order of stroke length. Essentially, for each full sketch, there can be as many \textit{category-epitome}s as the number of stroke orderings. In our analysis that follows, we keep the stroke ordering fixed over the set of categories. Details on stroke orderings and their effect on \textit{category-epitome}s that result can be found in Sections $4,5$ of ~\cite{eotd}. 

The procedure described in Section \ref{sec:overview} is used to generate semantic-part word clouds for the $13$ object categories we have chosen. Figure \ref{fig:epitome-wordle} shows the word clouds for the \fauxsc{Length}-based stroke ordering. Remember that the size of the semantic-part's name indicates its dominance in the sparsified representations (\textit{epitomes}). Categories which exhibit one or two dominant parts (e.g. \texttt{horse, dog, potted plant}) indicate that such parts are consistently present in most of the \textit{epitome}s. This, in turn, suggests a consistency in which sketches of the category are drawn. word clouds of categories with more variety in depictions (e.g. \texttt{airplane, person walking}) tend to contain many parts whose names are similar in size. Another interesting trend exists across semantically related categories. For instance, `leg' is found to be the common defining signature part for the animal categories (\texttt{cow, dog, horse, person walking, sheep}). Similarly, for the vehicular categories (\texttt{car, bus, bicycle, train}), `wheel' is a dominant part and for the flying categories (\texttt{airplane, bird}), `wing' is a dominant part. 

The trends mentioned above can also be seen for the \fauxsc{Temporal} stroke ordering (see Figure \ref{fig:epitome-wordle-temporal}). We can observe that the part importance trends are fairly same for each category across the stroke orderings. The epitomes created under \fauxsc{Temporal} stroke ordering scheme tend to contain the sequence of strokes added towards the beginning. Since the part importance trends for   \fauxsc{Temporal} are not very dissimilar from the \fauxsc{Length}-based ordering, this suggests, somewhat counter-intuitively, that people do not necessarily draw the ``signature" parts of a sketch first. The \fauxsc{Alternate} stroke ordering consists of an alternating combination of longest strokes and decorative strokes (temporally reversed order). However, even in this case, the essential dominance of ``signature" parts remains more or less unchanged across the categories (Figure \ref{fig:epitome-wordle-alternate}). These results across the stroke ordering schemes suggest that the ``signature" semantic parts live up to their name -- they capture the discriminative structural elements of the category and are invariant to the manner in which sketch strokes are considered in the process of \textit{epitome} construction. 

A more traditional, tabular version of the part word clouds with numerical values for various stroke orderings can be viewed in Tables \ref{tab:pi-temporal}, \ref{tab:pi-length} and \ref{tab:pi-alternate}.

\section{Discussion and Future Work}
\label{sec:futurework}

In this paper, we have presented a novel framework for analyzing the structural characteristics of \textit{category-epitome}s. We have shown that semantic-part annotations of sketches can be utilized to gain an intuitive understanding of category-level and sketch-stroke-ordering level structural trends in \textit{category-epitome}s. The database of part-annotated sketches of object categories is another significant contribution of our work since we can now simultaneously analyze relationships with photographic image counterparts at a semantic-part level. Finally, the word cloud based analysis we have presented is quite general and can be applied to any spatial visual object representation wherein the part labelings have been provided. 

At present, we have confined our analysis to the sketch database of Eitz et al.~\cite{eitz}. To examine the generalizability of our approach and results, it would be interesting to apply it to the part-segmented sketch database of Huang et al.~\cite{hongbofu}. Another possible extension would be to apply the sketch-part segmentation method suggested by the aforementioned authors for the entire set of categories (instead of the $13$ we have chosen) from the database of Eitz et al.~\cite{eitz}.

\newpage

\appendix

\section{Pseudo-code}
\label{sec:pseudocode}

\begin{minipage}{0.49\textwidth}

\begin{algorithm}[H] 
{\scriptsize 
\caption{Algorithm to determine contribution of structural part strokes to category-epitome}
  \begin{algorithmic}[1]
	\Procedure{GetEpiPW}{$catName,strokeOrd$}
	\LineComment{\textcolor[rgb]{0.2,0.8,0.2}{$catName$ - Name of the category (e.g. \texttt{bicycle})}} 
	\LineComment{\textcolor[rgb]{0.2,0.8,0.2}{$strokeOrd$ - Stroke sequence ordering (\textsc{Temporal,Length,Alternate})}} 		
	\item[]
	\LineComment{\textcolor[rgb]{0.2,0.8,0.2}{Get list of structural parts for the category}}	
	\State \textcolor[rgb]{0,0,1}{$P \gets $\Call{GetListOfParts}{$catName$}}	
	\LineComment{\textcolor[rgb]{0.2,0.8,0.2}{Get part annotations $U$ for sketches in the category}}	
	\State \textcolor[rgb]{0,0,1}{$U \gets $\Call{GetUserAnnotations}{$catName$}}	
	\LineComment{\textcolor[rgb]{0.2,0.8,0.2}{Initialize accumulator for weight of each part across epitomes of the category}}
	\State \textcolor[rgb]{0,0,1}{$epi\_wts\_freq \gets zeros(length(P),1)$}	
	\item[]
	\LineComment{\textcolor[rgb]{0.2,0.8,0.2}{For each correctly classified test image in the category}}	
	\For{\textcolor[rgb]{0,0,1}{$f = 1 : length(U)$}}	
	\State \textcolor[rgb]{0,0,1}{$[I,epi\_I] \gets $\Call{GetImages}{$U[f],strokeOrd$}}
	\LineComment{\textcolor[rgb]{0.2,0.8,0.2}{I = full sketch, epi\_I = corresponding category-epitome}}
	\LineComment{\textcolor[rgb]{0.2,0.8,0.2}{Get weights of parts as present in category-epitome epi\_I}}
	\State \textcolor[rgb]{1,0,0}{$epi\_wts\_f \gets $\Call{GetPartStats}{$U[f],I,epi\_I,P$}}
	\LineComment{\textcolor[rgb]{0.2,0.8,0.2}{Update accumulator for part weights with contribution from current epitome}}
	\State \textcolor[rgb]{0,0,1}{$epi\_wts\_freq \gets epi\_wts\_freq + epi\_wts\_f$}     	
	\EndFor
	\item[]
	\LineComment{\textcolor[rgb]{0.2,0.8,0.2}{Normalize the part weights}}	
	\State \textcolor[rgb]{0,0,1}{$pt\_wts\_freq \gets \frac{epi\_wts\_freq}{max(epi\_wts\_freq)}$}     
	\LineComment{\textcolor[rgb]{0.2,0.8,0.2}{Sort the part weights in decreasing order}}	
	\State \textcolor[rgb]{0,0,1}{$[sorted\_wts,sorted\_pt\_ids] \gets $\Call{Sort}{$pt\_wts\_freq$}}     
	\LineComment{\textcolor[rgb]{0.2,0.8,0.2}{return the list of parts and their weights in sorted order}}	
	\State \textcolor[rgb]{0,0,1}{$sorted\_part\_list \gets P(sorted\_pt\_ids)$}
	\State \Return \textcolor[rgb]{0,0,1}{$[sorted\_part\_list,sorted\_wts]$}    	
	\EndProcedure
	\end{algorithmic}
	\label{alg:one}
	}
\end{algorithm}

\end{minipage}

\begin{minipage}{0.49\textwidth}
\begin{algorithm}[H] 
{\scriptsize 
\caption{Analyzes stroke information of an epitome using the part annotations of corresponding original sketch.
 Obtains a listing of object parts which are prominent in the epitome and their relative importance.}
  \begin{algorithmic}[1]
	\Procedure{GetPartStats}{$U,I,epi\_I,P$}
	\LineComment{\textcolor[rgb]{0.2,0.8,0.2}{$U$ - part annotations for the sketch}} 	
	\LineComment{\textcolor[rgb]{0.2,0.8,0.2}{$I$ - Full sketch image}} 		
	\LineComment{\textcolor[rgb]{0.2,0.8,0.2}{$epi\_I$ - Category-epitome image}} 		
	\LineComment{\textcolor[rgb]{0.2,0.8,0.2}{$P$ - list of structural parts for the category}} 		
	\item[]
	\LineComment{\textcolor[rgb]{0.2,0.8,0.2}{Get frequency count of each part in full sketch}}	
	\State \textcolor[rgb]{0,0,1}{$parts\_list\_freq \gets $\Call{GetPartCount}{$U,P$}}	
	\LineComment{\textcolor[rgb]{0.2,0.8,0.2}{Get candidate contours}}	
	\State \textcolor[rgb]{1,0,0}{$cand\_contour\_id \gets $\Call{GetCandidateContours}{$U,epi\_I,P$}}		
	\LineComment{\textcolor[rgb]{0.2,0.8,0.2}{Get frequency count of parts w.r.t candidate contour list}}	
	\State \textcolor[rgb]{0,0,1}{$epi\_parts\_list\_freq \gets \Call{GetPartCount}{$U,P,cand\_contour\_id$}$}			
	\LineComment{\textcolor[rgb]{0.2,0.8,0.2}{Normalize for multiple occurrences of same part}}	
	\State \textcolor[rgb]{0,0,1}{$p\_im \gets epi\_parts\_list\_freq ./ parts\_list\_freq$} 
	\State \textcolor[rgb]{0,0,1}{$epi\_wts\_I \gets zeros(1,length(P)$}
	\For{\textcolor[rgb]{0,0,1}{$p = 1 : length(cand\_contour\_id)$}}
		\LineComment{\textcolor[rgb]{0.2,0.8,0.2}{Count stroke pixels of epitome ``hugging" candidate contour's boundary}}	
		\State \textcolor[rgb]{1,0,0}{$n\_valid\_epi \gets $\Call{CtValidMatch}{$I\_epi,U,cand\_contour\_id[p]$}}		
	\LineComment{\textcolor[rgb]{0.2,0.8,0.2}{Count stroke pixels of fullsketch ``hugging" candidate contour's boundary}}	
	\State \textcolor[rgb]{1,0,0}{$n\_valid\_full \gets $\Call{CtValidMatch}{$I,U,cand\_contour\_id[p]$}}		
	\LineComment{\textcolor[rgb]{0.2,0.8,0.2}{Get the importance of the corresponding `part' }}	
	\State \textcolor[rgb]{0,0,1}{$w\_part \gets \frac{n\_valid\_epi}{n\_valid\_full}$}
	\State \textcolor[rgb]{0,0,1}{$epi\_part\_name \gets U[cand\_contour\_id[p]].part\_name$}			
	\For{\textcolor[rgb]{0,0,1}{$s = 1 : length(P)$}}
	\If{\textcolor[rgb]{0,0,1}{$P[s] == epi\_part\_name$}} 
	\State \textcolor[rgb]{0,0,1}{$epi\_wts\_I[s] \gets epi\_wts\_I[s] + (p\_im[s] \times w\_part)$} 
	\State \textcolor[rgb]{0,0,0}{\textbf{break}}	
	\EndIf
	\EndFor
	\EndFor	
	\State \Return \textcolor[rgb]{0,0,1}{$epi\_wts\_I$}
	\EndProcedure
	\end{algorithmic}
	\label{alg:two}
	}
\end{algorithm}
\end{minipage}

\begin{minipage}{0.49\textwidth}
\begin{algorithm}[H] 
{\scriptsize 
\caption{Gets list of candidate contours}
  \begin{algorithmic}[1]
	\Procedure{GetCandidateContours}{$U,I,epi\_I,P$}
	\LineComment{\textcolor[rgb]{0.2,0.8,0.2}{$I$ - Full sketch image}} 		
	\LineComment{\textcolor[rgb]{0.2,0.8,0.2}{$U$ - part annotation array for full sketch image}}
	\LineComment{\textcolor[rgb]{0.2,0.8,0.2}{$epi\_I$ - Category-epitome image}} 		
	\LineComment{\textcolor[rgb]{0.2,0.8,0.2}{$P$ - list of structural parts for the category}} 		
	\item[]
	\For{\textcolor[rgb]{0,0,1}{$p = 1 : length(U)$}}
	\LineComment{\textcolor[rgb]{0.2,0.8,0.2}{Get 2D part contour from user annotation}}	
	\State \textcolor[rgb]{0,0,1}{$[xd,yd] \gets $\Call{GetPartContour}{$U[p]$}}	
	\LineComment{\textcolor[rgb]{0.2,0.8,0.2}{Get count of stroke pixels within the part contour from the full sketch image}}	
	\State \textcolor[rgb]{0,0,1}{$num\_stroke\_pixels\_orig \gets $\Call{CountPixels}{$xd,yd,I$}}		
	\LineComment{\textcolor[rgb]{0.2,0.8,0.2}{Get count of stroke pixels within the part contour from epitome}}	
	\State \textcolor[rgb]{0,0,1}{$num\_stroke\_pixels\_epi \gets $\Call{CountPixels}{$xd,yd,epi\_I$}}		
	\LineComment{\textcolor[rgb]{0.2,0.8,0.2}{If number of stroke pixels within part contour from the epitome is greater than a threshold, add the part as a candidate part}}	
	\State \textcolor[rgb]{0,0,1}{$part\_membership\_ratio \gets \frac{num\_stroke\_pixels\_epi}{num\_stroke\_pixels\_orig}$}	
	\If{\textcolor[rgb]{0,0,1}{$part\_membership\_ratio > \epsilon$}} 
	\State \textcolor[rgb]{0,0,1}{cand\_contour\_ids.\textsc{Insert}(p) } 	
	\EndIf
	\EndFor
	\State \Return \textcolor[rgb]{0,0,1}{$cand\_contour\_ids$}
	\EndProcedure
	\end{algorithmic}
	\label{alg:three}
	}
\end{algorithm}
\end{minipage}

\begin{minipage}{0.49\textwidth}
\begin{algorithm}[H] 
{\scriptsize 
\caption{Gets list of candidate parts (which potentially contribute) in the epitome}
  \begin{algorithmic}[1]	
	\Procedure{AnalyzeParts}{$can\_p\_id,U,I,epi\_I,P,p\_i$}	
	\LineComment{\textcolor[rgb]{0.2,0.8,0.2}{$can\_p\_id[p]$ - candidate part}} 		
	\LineComment{\textcolor[rgb]{0.2,0.8,0.2}{$U$ -  part annotations for the sketch}} 	
	\LineComment{\textcolor[rgb]{0.2,0.8,0.2}{$I$ - Full sketch image}} 		
	\LineComment{\textcolor[rgb]{0.2,0.8,0.2}{$epi\_I$ - Category-epitome image}} 
	\LineComment{\textcolor[rgb]{0.2,0.8,0.2}{$P$ - list of structural parts for the category}} 		
	\LineComment{\textcolor[rgb]{0.2,0.8,0.2}{$p\_i$ - Part-wise importance factor}}
	\item[]
	\LineComment{\textcolor[rgb]{0.2,0.8,0.2}{Count stroke pixels of epitome which ``hug" the candidate part's boundary}}	
	\State \textcolor[rgb]{1,0,0}{$n\_valid\_epi \gets $\Call{CtValidMatch}{$I\_epi,U,can\_p\_id$}}		
	\LineComment{\textcolor[rgb]{0.2,0.8,0.2}{Count stroke pixels of full sketch which ``hug" the candidate part's boundary}}	
	\State \textcolor[rgb]{1,0,0}{$n\_valid\_full \gets $\Call{CtValidMatch}{$I,U,can\_p\_id$}}		
	\LineComment{\textcolor[rgb]{0.2,0.8,0.2}{Get the importance of this candidate part}}	
	\State \textcolor[rgb]{0,0,1}{$w\_component \gets \frac{n\_valid\_epi}{n\_valid\_full}$}
	\State \textcolor[rgb]{0,0,1}{$epi\_part\_name \gets U[f].part\_name$}
	\State \textcolor[rgb]{0,0,1}{$epi\_weights\_I \gets zeros(length(P),1)$}
	\LineComment{\textcolor[rgb]{0.2,0.8,0.2}{Account for multiple occurences of same part in the epitome}}	
	\For{\textcolor[rgb]{0,0,1}{$s = 1 : length(P)$}}
		\If{\textcolor[rgb]{0,0,1}{$P[s] == epi\_part\_name$}} 
			\State \textcolor[rgb]{0,0,1}{$epi\_weights\_I[s] \gets epi\_weights\_I[s] + p\_i[s] \times w\_component[s]$} 	
	  \EndIf	
	\EndFor		
	\State \Return \textcolor[rgb]{0,0,1}{$epi\_weights\_I$}
	\EndProcedure
	\end{algorithmic}
	\label{alg:four}
	}
\end{algorithm}
\end{minipage}

\begin{minipage}{0.49\textwidth}
\begin{algorithm}[H] 
{\scriptsize 
\caption{Counts the number of stroke pixels of image that lie ``close" to candidate part's contour}
  \begin{algorithmic}[1]	
	\Procedure{CtValidMatch}{$I,U,cand\_part\_id$}
	\LineComment{\textcolor[rgb]{0.2,0.8,0.2}{$I$ - Sketch image (full or epitome)}}
	\LineComment{\textcolor[rgb]{0.2,0.8,0.2}{$U$ - array of part annotations for sketches in the category}} 
	\LineComment{\textcolor[rgb]{0.2,0.8,0.2}{$cand\_part\_id$ - Index into array $U$}} 		
	\item[]
	\LineComment{\textcolor[rgb]{0.2,0.8,0.2}{Get 2D part contour from user annotation}}	
	\State \textcolor[rgb]{0,0,1}{$[xd,yd] \gets U.$\Call{GetPartContour}{$cand\_part\_id$}}
	\LineComment{\textcolor[rgb]{0.2,0.8,0.2}{Get stroke pixels from image which lie inside candidate part's mask}}	
	\State \textcolor[rgb]{0,0,1}{$[x\_lc,y\_lc] \gets $\Call{GetStrokePixels}{$xd,yd,I$}}		
	\LineComment{\textcolor[rgb]{0.2,0.8,0.2}{For each stroke pixel p from image which lies inside candidate part's mask}}	
	\LineComment{\textcolor[rgb]{0.2,0.8,0.2}{   Find the nearest pixel p' on the candidate part boundary}}
	\State \textcolor[rgb]{0,0,1}{$[min\_index,min\_dist] \gets $}
	\State \textcolor[rgb]{0,0,1}{$ $\Call{GetNearest}{$[xd\hspace{3mm}yd],[x\_lc\hspace{3mm}y\_lc],'K',1$}}		
	\LineComment{\textcolor[rgb]{0.2,0.8,0.2}{Retain the matches whose distance is less than a threshold}}
	\State \textcolor[rgb]{0,0,1}{$filtered\_ids \gets $\Call{FilterList}{$min\_dist,THRESH$}}		
	\State \textcolor[rgb]{0,0,1}{$[xd\_f,yd\_f] \gets $\Call{Index}{$xd,yd,min\_index,filtered\_ids$}}		
	\Comment{\textcolor[rgb]{0.2,0.8,0.2}{[xd\_f yd\_f] - Filtered matching boundary points}}
	\State \textcolor[rgb]{0,0,1}{$num\_valid\_matches \gets $\Call{Length}{$xd\_f,yd\_f,'unique'$}}
	\State \Return \textcolor[rgb]{0,0,1}{$num\_valid\_matches$}
  \EndProcedure
	\end{algorithmic}
	\label{alg:five}
	}
\end{algorithm}
\end{minipage}

\clearpage

\begin{table*}[!ht]
\begin{tabularx}{\textwidth}{|l|X|}
\hline
Category & Epitome part-list and weights (\textsc{temporal}) \\
\hline
\hline
airplane &  window (1.000),  wing (0.373),  fuselage (0.190),  vertical stabilizer (0.183),  wind shield (0.159),  horizontal stabilizer (0.151),  engine (0.095),  door (0.048),  nose (0.008) \\
 \hline
bicycle &  spoke (1.000),  frame (0.441),  wheel (0.304),  handlebars (0.147),  seat (0.127),  pedal (0.093),  chain (0.088) \\
 \hline
bus &  window (1.000),  wheel (0.421),  body (0.220),  windshield (0.101),  headlight (0.094),  door (0.088),  steering (0.044),  roof (0.038) \\
 \hline
car (sedan) &  wheel (1.000),  window (0.963),  frame (0.481),  door (0.315),  headlight  (0.259),  windshield (0.148),  bumper (0.111),  bonnet (0.074),  seat (0.056),  steering (0.037),  radiator grille (0.037) \\
 \hline
cat &  whiskers (1.000),  paw (0.531),  eye (0.449),  ear (0.449),  leg  (0.245),  nose (0.224),  tail (0.204),  mouth (0.143) \\
 \hline
cow &  leg (1.000),  ear (0.481),  eye (0.462),  patch (0.327),  horn (0.308),  tail (0.308),  udder (0.269),  mouth (0.231),  nose (0.173) \\
 \hline
dog &  leg (1.000),  eye (0.405),  ear (0.333),  nose (0.286),  body  (0.286),  head (0.286),  tail (0.262),  mouth (0.190) \\
 \hline
flying bird &  wing (1.000),  beak (0.500),  head  (0.500),  body (0.500),  tail (0.500),  eye (0.455),  leg (0.091) \\
 \hline
horse &  leg (1.000),  hoofs (0.310),  eye (0.264),  head (0.264),  tail (0.264),  mane (0.230),  mouth (0.138),  nose (0.138),  body (0.069) \\
 \hline
person walking &  leg (1.000),  hand (0.940),  foot (0.860),  head (0.520),  eye (0.480),  mouth (0.240),  chest (0.240),  hair (0.140),  nose (0.100) \\
 \hline
potted plant &  leaf (1.000),  stem (0.382),  pot (0.224),  flower (0.127) \\
 \hline
sheep &  leg (1.000),  eye  (0.359),  ear (0.321),  mouth (0.269),  body  (0.269),  tail (0.167),  nose (0.000) \\
 \hline
train &  wheel (1.000),  window (0.578),  coach (0.311),  engine (0.156),  chimney (0.139),  smoke (0.128),  coupler	 (0.122),  track (0.117),  front (0.061),  door (0.028) \\
 \hline
\end{tabularx}
\caption{Category-wise part-importances for \textsc{temporal} stroke ordering. Part-importances are listed in decreasing order of importance relative to the most dominant part (shown with weight 1)}
\label{tab:pi-temporal}
\end{table*}

\begin{table*}[!htbp]
\begin{tabularx}{\textwidth}{|l|X|}
\hline
Category & Epitome part-list and weights (\textsc{length}) \\
\hline
\hline
airplane &  wing (1.000),  window (0.692),  fuselage (0.522),  vertical stabilizer (0.423),  wind shield (0.340),  horizontal stabilizer (0.223),  engine (0.121),  door (0.067),  nose (0.023) \\
 \hline
bicycle &  frame (1.000),  wheel (0.936),  spoke (0.727),  seat (0.239),  chain (0.189),  handlebars (0.149),  pedal (0.124) \\
 \hline
bus &  window (1.000),  wheel (0.629),  body (0.346),  windshield (0.150),  door (0.132),  roof (0.063),  steering (0.045),  headlight (0.044) \\
 \hline
car (sedan) &  wheel (1.000),  window (0.767),  frame (0.619),  door (0.254),  headlight  (0.162),  windshield (0.131),  bonnet (0.106),  radiator grille (0.040),  bumper (0.032),  seat (0.027),  steering (0.024) \\
 \hline
cat &  whiskers (1.000),  paw (0.867),  ear (0.717),  leg  (0.394),  tail (0.334),  eye (0.317),  nose (0.232),  mouth (0.066) \\
 \hline
cow &  leg (1.000),  ear (0.327),  patch (0.299),  tail (0.261),  horn (0.258),  udder (0.220),  mouth (0.158),  nose (0.114),  eye (0.110) \\
 \hline
dog &  leg (1.000),  head (0.306),  body  (0.297),  ear (0.290),  tail (0.287),  nose (0.173),  mouth (0.166),  eye (0.104) \\
 \hline
flying bird &  wing (1.000),  tail (0.549),  body (0.479),  head  (0.314),  beak (0.225),  leg (0.011),  eye (0.000) \\
 \hline
horse &  leg (1.000),  tail (0.295),  mane (0.237),  hoofs (0.236),  head (0.232),  eye (0.097),  body (0.072),  nose (0.059),  mouth (0.058) \\
 \hline
person walking &  leg (1.000),  foot (0.751),  hand (0.638),  head (0.507),  chest (0.233),  hair (0.094),  mouth (0.037),  nose (0.030),  eye (0.005) \\
 \hline
potted plant &  leaf (1.000),  stem (0.466),  pot (0.359),  flower (0.171) \\
 \hline
sheep &  leg (1.000),  body  (0.401),  mouth (0.272),  ear (0.259),  tail (0.171),  eye  (0.035),  nose (0.000) \\
 \hline
train &  wheel (1.000),  coach (0.399),  window (0.318),  engine (0.195),  track (0.141),  chimney (0.140),  coupler	 (0.119),  smoke (0.095),  front (0.078),  door (0.013) \\
 \hline
\end{tabularx}
\caption{Category-wise part-importances for \textsc{length} stroke ordering. Part-importances are listed in decreasing order of importance relative to the most dominant part (shown with weight 1)}
\label{tab:pi-length}
\end{table*}

\begin{table*}[!htbp]
\begin{tabularx}{\textwidth}{|l|X|}
\hline
Category & Epitome part-list and weights (\textsc{alternate}) \\
\hline
\hline
airplane &  wing (1.000),  window (0.928),  fuselage (0.599),  vertical stabilizer (0.471),  wind shield (0.334),  engine (0.256),  horizontal stabilizer (0.193),  door (0.054),  nose (0.027) \\
 \hline
bicycle &  wheel (1.000),  spoke (0.960),  frame (0.933),  seat (0.263),  handlebars (0.260),  chain (0.207),  pedal (0.097) \\
 \hline
bus &  window (1.000),  wheel (0.567),  body (0.316),  windshield (0.130),  door (0.128),  headlight (0.065),  roof (0.056),  steering (0.054) \\
 \hline
car (sedan) &  wheel (1.000),  window (0.935),  frame (0.651),  headlight  (0.324),  door (0.281),  windshield (0.142),  bonnet (0.115),  radiator grille (0.059),  bumper (0.057),  steering (0.022),  seat (0.010) \\
 \hline
cat &  whiskers (1.000),  paw (0.992),  ear (0.817),  leg  (0.447),  tail (0.389),  eye (0.336),  nose (0.229),  mouth (0.183) \\
 \hline
cow &  leg (1.000),  ear (0.373),  horn (0.308),  patch (0.304),  tail (0.263),  eye (0.257),  udder (0.236),  mouth (0.174),  nose (0.123) \\
 \hline
dog &  leg (1.000),  head (0.328),  body  (0.324),  tail (0.304),  eye (0.295),  ear (0.294),  nose (0.273),  mouth (0.193) \\
 \hline
flying bird &  wing (1.000),  tail (0.551),  body (0.514),  head  (0.387),  beak (0.295),  eye (0.102),  leg (0.054) \\
 \hline
horse &  leg (1.000),  tail (0.283),  hoofs (0.259),  head (0.242),  mane (0.211),  eye (0.196),  mouth (0.112),  nose (0.096),  body (0.079) \\
 \hline
person walking &  leg (1.000),  foot (0.839),  hand (0.821),  head (0.517),  chest (0.247),  mouth (0.136),  eye (0.125),  hair (0.118),  nose (0.072) \\
 \hline
potted plant &  leaf (1.000),  stem (0.388),  pot (0.345),  flower (0.145) \\
 \hline
sheep &  leg (1.000),  body  (0.440),  eye  (0.260),  mouth (0.251),  ear (0.214),  tail (0.195),  nose (0.000) \\
 \hline
train &  wheel (1.000),  window (0.388),  coach (0.369),  engine (0.176),  track (0.153),  chimney (0.123),  coupler	 (0.121),  smoke (0.074),  front (0.062),  door (0.011) \\
 \hline
\end{tabularx}
\caption{Category-wise part-importances for \textsc{alternate} stroke ordering. Part-importances are listed in decreasing order of importance relative to the most dominant part (shown with weight 1)}
\label{tab:pi-alternate}
\end{table*}

\clearpage

{\small
\bibliographystyle{ieee}
\bibliography{egbib}
}

\end{document}